\pdfoutput=1

\documentclass[review,11pt,authoryear]{elsarticle}

\usepackage{amsmath}
\usepackage{amssymb}
\usepackage{amsfonts}
\usepackage{amsthm}
\usepackage{natbib}
\usepackage{enumitem}
\usepackage{graphicx}
\usepackage{tikz}
\usepackage{booktabs}
\usepackage{multirow}
\usepackage{placeins}
\usepackage{needspace}
\usepackage{xcolor}
\usepackage[hidelinks]{hyperref}

\newcommand{\revadd}[1]{{#1}}

\theoremstyle{plain}
\newtheorem{theorem}{Theorem}
\newtheorem{lemma}{Lemma}
\newtheorem{proposition}{Proposition}
\newtheorem{corollary}{Corollary}

\theoremstyle{definition}
\newtheorem{assumption}{Assumption}
\newtheorem{definition}{Definition}
\newtheorem{model}{Model}

\theoremstyle{remark}

\DeclareMathOperator*{\argmin}{arg\,min}
\DeclareMathOperator{\KL}{KL}
\DeclareMathOperator{\Var}{Var}
\DeclareMathOperator{\Cov}{Cov}

\newcommand{\E}{\mathbb{E}}
\newcommand{\Prob}{\mathbb{P}}
\newcommand{\R}{\mathbb{R}}
\newcommand{\Z}{\mathbb{Z}}

\newcommand{\filt}[1]{\mathcal{Y}_{#1}}
\newcommand{\volsq}{\sigma^{2}}
\newcommand{\volhat}[1]{\widehat{\sigma}^{2}_{#1}}
\newcommand{\mds}{V}
\newcommand{\ftrue}{f_{*}}
\newcommand{\fbase}{f_{0}}
\newcommand{\fpert}{f_{\theta}}
\newcommand{\Xtrue}[1]{X_{#1,*}}

\newcommand{\pstar}{p^{*}}
\newcommand{\qstar}{q^{*}}
\newcommand{\fdim}{p^{*}+q^{*}}

\newcommand{\Fclass}{\mathcal{F}}
\newcommand{\wnet}{\mathcal{W}}
\newcommand{\dnet}{\mathcal{D}}
\newcommand{\snet}{\mathcal{S}}
\newcommand{\Bmin}{B_{\min}}
\newcommand{\Bmax}{B_{\max}}
\newcommand{\Dcon}{D}
\newcommand{\hol}{\beta}
\newcommand{\mixcoef}[1]{\beta^{\mathrm{mix}}_{#1}}

\newcommand{\Ncell}{N_{\mathrm{cell}}}
\newcommand{\Cclass}{\mathcal{C}}

\newcommand{\mbump}{\psi}
\newcommand{\bumpj}[1]{\psi_{#1}}
\newcommand{\amp}{a_T}
\newcommand{\ampz}{a_{0}}
\newcommand{\hcube}{h}

\newcommand{\Emix}{\E_{\mathrm{mix}}}

\newcommand{\desfreq}[1]{\nu_{#1}}
\newcommand{\logodds}[1]{\Lambda_{#1}}
\newcommand{\postvar}[1]{\upsilon_{#1}}

\begin{document}

\begin{frontmatter}

\title{Nonparametric GARCH: An Iterated Neural Estimator}

\author[polyu]{Ruizhi Deng}
\author[polyu]{Guohao Shen}
\author[cityu]{Ngai Hang Chan}

\nonumnote{\hspace*{-1.8em}\raggedright
\mbox{Ruizhi~Deng:~\href{mailto:ruizhi.deng@connect.polyu.hk}{ruizhi.deng@connect.polyu.hk}},
\mbox{Guohao~Shen:~\href{mailto:guohao.shen@polyu.edu.hk}{guohao.shen@polyu.edu.hk}},
\mbox{Ngai~Hang~Chan:~\href{mailto:nhchan@cityu.edu.hk}{nhchan@cityu.edu.hk}}.}

\affiliation[polyu]{organization={Department of Applied Mathematics,
            The Hong Kong Polytechnic University}}
\affiliation[cityu]{organization={Department of Biostatistics,
            City University of Hong Kong}}

\begin{abstract}
We study nonparametric estimation and forecasting for GARCH-type processes and
propose an iterated neural estimator that alternates between estimating the
recursion map and updating the unobserved conditional variances.  Under mild
conditions, we establish finite-sample bounds for the time-averaged in-sample
variance-path error and show that the estimator is minimax rate-optimal up to
logarithmic factors.  Under stationarity and mixing conditions, we establish
the same rate for recursion-map estimation and derive finite-sample bounds for
out-of-sample variance forecasts and expected excess Gaussian quasi-likelihood
(QLIKE) loss.  The convergence rates depend on the true lag orders and remain
unchanged when the fitted orders are overspecified.  Simulation
studies illustrate the effectiveness of our estimators under nonlinear and
time-varying volatility dynamics.
Real-data applications suggest that our neural methods perform
competitively and can incorporate option-price and high-frequency information.
\end{abstract}


\end{frontmatter}

\section{Introduction}\label{sec:intro}

Conditional volatility is central to risk management, asset pricing, and
portfolio choice.  Accurate volatility
forecasts are important for econometric analysis and financial
decision making
\citep{engle2004risk,poon2003forecasting,andersen2006volatility}.
Autoregressive conditional heteroskedasticity (ARCH) models specify
conditional variance as a parametric function of past squared returns
\citep{Engle1982}.  Generalized autoregressive conditional heteroskedasticity
(GARCH) models also include past conditional variances
\citep{Bollerslev1986}.  These models and their extensions can describe how
return shocks affect volatility over several periods and how this effect
differs between positive and negative returns
\citep{EngleBollerslev1986,Nelson1991,GlostenJagannathanRunkle1993}.
\revadd{Although these models use relatively few parameters and are widely
applied, their accuracy depends on the chosen functional form
\citep{BollerslevEngleNelson1994,Shephard1996}.}

Nonparametric GARCH models offer greater flexibility by modeling the
conditional variance as a smooth function of past returns and past
conditional variances
\citep{masry1995nonparametric,fan1998efficient,linton2005estimating}.
However, without realized measures constructed from intraday
returns \citep{andersen2003modeling,hansen2012realized}, estimation is difficult
because both the variance function and its lagged variance inputs must be
estimated from the observed returns.
\citet{buhlmann2002algorithm} address these difficulties by proposing an
iterative algorithm that alternates between estimating the variance function
and updating the latent variance path.  \revadd{At each iteration, a
local-polynomial or a kernel regression is used to estimate the unknown function
\citep{HardleTsybakov1997}.} These methods can handle one or a few inputs, but
their accuracy deteriorates as the number of relevant lags increases.

\revadd{Deep neural networks have shown strong performance in nonparametric
regression with many inputs \citep{bauer2019deep,farrell2021deep}, which can
estimate nonlinear recursions in nonparametric GARCH models.} \revadd{A related
empirical literature uses neural networks and machine-learning methods to
forecast volatility
\citep{kim2018forecasting,bucci2020realized,christensen2023machine}.} These
studies focus on forecasting, while existing statistical theory establishes
approximation and regression properties for neural networks under independent
observations
\citep{yarotsky2018optimal,petersen2018optimal,lu2020deep,
bartlett2019nearly,schmidt2020nonparametric,jiao2023deep}.  Recent work develops related
regression theory for temporally dependent data
\citep{kuznetsov2015learning,kengne2023excess,kurisu2025adaptive} and for
first-order nonparametric GARCH models \citep{meister2016statistical}.
However, the deep-network results assume that the regressors are observed and
therefore do not directly provide a theory for nonparametric GARCH estimation.
\revadd{\citet{buhlmann2002algorithm} derive an error bound for their
iterative algorithm under conditions on the estimated variance inputs,
which are difficult to verify, so the proposed bound has not provided a
complete convergence theory for nonparametric GARCH estimation.}

In this paper, we propose an iterated neural estimator for nonparametric GARCH
models with general return and variance lag orders and establish finite-sample
guarantees for the estimator and its forecasts.  We show that the estimator
adapts to the true lag orders when the fitted lag orders are overspecified.
We also introduce a weighted extension that places more emphasis on recent
observations when volatility dynamics change over time.
Simulation studies illustrate the effectiveness of
our estimators.
\revadd{The main contributions are as follows.}

\begin{itemize}
\item[1.] We propose an iterated neural estimator for nonparametric GARCH
models with general lag orders.  It accommodates complex nonlinear
volatility dynamics using standard neural-network regressions that are
straightforward to implement.  We also develop a weighted extension for
volatility dynamics that change over time.

\item[2.] We establish finite-sample bounds for the in-sample variance path,
the recursion map, and out-of-sample forecasts.  The in-sample
path attains the classical nonparametric regression rate up to logarithmic
factors \citep{stone1982optimal}, with a matching minimax lower bound when the
target function has a smoothness index no less than one.  Under additional stationarity and regularity
conditions, we show that the recursion map and a one-step forecast after a
logarithmic updating period can attain the same rate.  We also prove that the
estimator retains the same convergence rate
when the fitted lag orders are overspecified.  To our knowledge,
these are the first complete finite-sample convergence guarantees for an
implementable iterative nonparametric GARCH estimator with general return and
variance lag orders.

\item[3.] We conduct numerical experiments with out-of-sample
forecast evaluation across linear, nonlinear, and time-varying designs.  The
results show that our estimator performs consistently well in most designs and
outperforms the benchmark estimators, particularly under nonlinear and complex
volatility dynamics.  The estimator is also insensitive to
lag-order overspecification, while the weighted extensions adapt better to
gradual changes in volatility dynamics.  Real-data applications
further illustrate that neural methods can incorporate option-price and
high-frequency information.
\end{itemize}

The rest of the paper is organized as follows.
Section~\ref{sec:nongarch} introduces the nonparametric GARCH model, its
regression formulation, and the iterated neural estimator.
Section~\ref{sec:main-results} gives the in-sample and
out-of-sample bounds, recursion-map recovery, lag-order adaptivity, and the
lower bound.  Section~\ref{sec:weighted-extension} gives a concise weighted
extension.
Section~\ref{sec_sim} reports the numerical study.
Section~\ref{sec:real-data} presents the real-data
applications.
Section~\ref{sec:conclusion} discusses the scope of the results and open
problems.  The supplementary material contains the proofs and additional
technical details.

\section{Model and Iterated Neural Estimator}\label{sec:nongarch}

A nonparametric GARCH model replaces the parametric recursion in ARCH and GARCH
models with an unspecified function that maps recent returns and variances to
the current conditional variance.  Next, we define the model and identify its
target function, formulate the estimation problem as a conditional-mean
regression, and introduce our neural estimator.

\subsection{The nonparametric GARCH model and its regression formulation}
\label{model:regression}

Let $\{Y_t:1\le t\le T\}$ be a return series observed over $T$ periods of the
process $(Y_t)_{t\in\Z}$.  Write
$\filt{t}=\sigma(Y_s:s\le t)$ for the sigma-algebra generated by the returns
up to time $t$, which is the information available at that time.  A
nonparametric GARCH model of orders $\pstar\ge1$ and $\qstar\ge0$ posits the
following conditional-variance recursion.

\begin{model}[Nonparametric GARCH]\label{model:struct}
The return series is generated by
\begin{equation}
\begin{aligned}
  Y_t\;&=\;\sigma_t\varepsilon_t,
  \qquad
  \volsq_t\;=\;\ftrue(\Xtrue{t}),\\
  \Xtrue{t}\;&=\;
  \bigl(Y_{t-1},\ldots,Y_{t-\pstar},
  \volsq_{t-1},\ldots,\volsq_{t-\qstar}\bigr)
  \in\R^{\fdim}.
\end{aligned}
\label{eq:struct}
\end{equation}
Here $\ftrue:\R^{\fdim}\to[0,\infty)$ is a measurable function with input
dimension $\pstar+\qstar$.  The innovations
$(\varepsilon_t)_{t\in\Z}$ are independent and identically distributed satisfying
$\E(\varepsilon_t)=0$, $\E(\varepsilon_t^2)=1$,
$\E(\varepsilon_t^4)<\infty$, and $\varepsilon_t$ is independent of
$\filt{t-1}$.  Here $\sigma_t$ is the volatility and
$\volsq_t=\Var(Y_t\mid\filt{t-1})$ is its conditional variance.
\end{model}

The conditional variance is a nonnegative function $\ftrue$ of the regressor
vector $\Xtrue{t}$, which contains the $\pstar$ most recent returns and the $\qstar$
most recent variances.  The values $\pstar$ and $\qstar$ give the largest
return lag and variance lag that can affect the current conditional variance.
When $\qstar=0$, the conditional variance depends only on lagged returns, and
Model~\ref{model:struct} reduces to a nonparametric ARCH model.  Parametric
ARCH and GARCH models arise when $\ftrue$ is affine in the squared return
inputs and variance lags
\citep{Engle1982,Bollerslev1986}.  Allowing $\ftrue$ to be a general
measurable function gives the nonparametric formulation
\citep{buhlmann2002algorithm}.

Our objective is to estimate both the recursion map $\ftrue$ and the latent
variance path $(\volsq_t)$ from the observed returns.  We first formulate this
problem as a regression.  Note that both
$\volsq_t=\ftrue(\Xtrue{t})$ and $\Xtrue{t}$ are
$\filt{t-1}$-measurable.  The independence and normalization of
$\varepsilon_t$ and the tower property give
\begin{equation}
\begin{aligned}
  \E(Y_t^2\mid\filt{t-1})
    &=\volsq_t\E(\varepsilon_t^2\mid\filt{t-1})
      =\volsq_t=\ftrue(\Xtrue{t}),\\
  \E(Y_t^2\mid\Xtrue{t})
    &=\ftrue(\Xtrue{t}).
\end{aligned}
\label{eq:cm}
\end{equation}
Therefore, $\ftrue$ is the population regression function of $Y_t^2$ given
$\Xtrue{t}$.  Define the regression error by
\begin{equation}
  \mds_t=Y_t^2-\volsq_t=\volsq_t(\varepsilon_t^2-1),
  \label{eq:mds}
\end{equation}
which satisfies $\E(\mds_t\mid\filt{t-1})=0$ by \eqref{eq:cm}.  The observable
squared return therefore satisfies the conditional-mean regression
\begin{equation}
  Y_t^2=\ftrue(\Xtrue{t})+\mds_t,
  \qquad
  \E(\mds_t\mid\filt{t-1})=0.
  \label{eq:regression}
\end{equation}
The regression error is heteroscedastic, with conditional variance
$\sigma_t^4\Var(\varepsilon_t^2)$.

When $\qstar=0$, $\Xtrue{t}$ contains only the observed return lags
$Y_{t-1},\ldots,Y_{t-\pstar}$.  We can therefore estimate $\ftrue$ directly
by a single regression.  When
$\qstar\ge1$, however, $\Xtrue{t}$ also contains the unobserved variance lags
$\volsq_{t-1},\ldots,\volsq_{t-\qstar}$.  Regression
\eqref{eq:regression} then cannot be implemented directly.  To address this
difficulty, we propose an iterated neural estimator in the next subsection.

\subsection{The iterated neural estimator}
\label{sec:algo}

Our estimator alternates between estimating the recursion map by least squares
over a function class and updating the latent variance path.
For fitted lag orders $p$ and $q$, let $r_0=\max(p,q)$ and write
$Y_{t-1:t-p}=(Y_{t-1},\ldots,Y_{t-p})$.  We also write
$\widehat\sigma^2_{t-1:t-q,m}=(\volhat{t-1,m},\ldots,\volhat{t-q,m})$
for the lagged variance estimates at iteration $m$.
Estimation proceeds in three steps.

\begin{enumerate}[leftmargin=2.4em,label=\textnormal{(\arabic*)}]
\item\textbf{Initialize.}
Set $\volhat{t,0}=\widehat c$ for $t=1,\ldots,T$ and retain
$\volhat{t,m}=\widehat c$ for $t\leq r_0$ at every subsequent iteration.  The
positive constant $\widehat c$ may be specified in advance or estimated from
the data, for example by maximum likelihood under a preliminary parametric
model.

\item\textbf{Estimate and update.}
Choose the number of iterations $M\geq1$ and a function class
$\Fclass$.
At each iteration $m=1,\ldots,M$, define the least-squares estimator
\begin{equation}
 \widehat f_m\in\argmin_{f\in\Fclass}
 \frac1{T-r_0}\sum_{t=r_0+1}^{T}
 \left[
 Y_t^2-f\left(Y_{t-1:t-p},\volhat{t-1:t-q,m-1}\right)
 \right]^2,
 \label{eq:apt-erm}
\end{equation}
where the squared returns $Y_t^2$ are observed and the variance lags are
estimated at iteration $m-1$.  For $t=r_0+1,\ldots,T$, use $\widehat f_m$ to
update the remaining variance estimates by
\begin{equation}
 \volhat{t,m}=\widehat f_m
 \left(Y_{t-1:t-p},\volhat{t-1:t-q,m-1}\right).
\label{eq:apt-update}
\end{equation}
Repeat the estimation and variance-update steps through iteration $M$.

\item\textbf{Return the estimates.}
Return $\widehat f_M$ as the estimator of the recursion map $\ftrue$ and
$(\volhat{t,M})_{t=r_0+1}^{T}$ as the estimated in-sample variance path.
\end{enumerate}

For forecasting, we first evaluate $\widehat f_M$ at $Y_{T:T-p+1}$ and
$\widehat\sigma^2_{T:T-q+1,M}$ to obtain the first post-estimation variance
forecast $\widehat\sigma^2_{T+1\mid T}$.  As each new return becomes available, we apply
the same update recursively to generate the next one-step-ahead variance forecast,
\begin{equation}
 \widehat\sigma^2_{t\mid T}
 =\widehat f_M
 \left(Y_{t-1:t-p},
 \widehat\sigma^2_{t-1:t-q\mid T}\right),
 \qquad t>T,
 \label{eq:apt-estimate}
\end{equation}
where $\widehat\sigma^2_{t-1:t-q\mid T}$ contains the $q$
in-sample variance estimates or forecasts.

\subsection{The deep neural network class}
\label{sec:netclass}

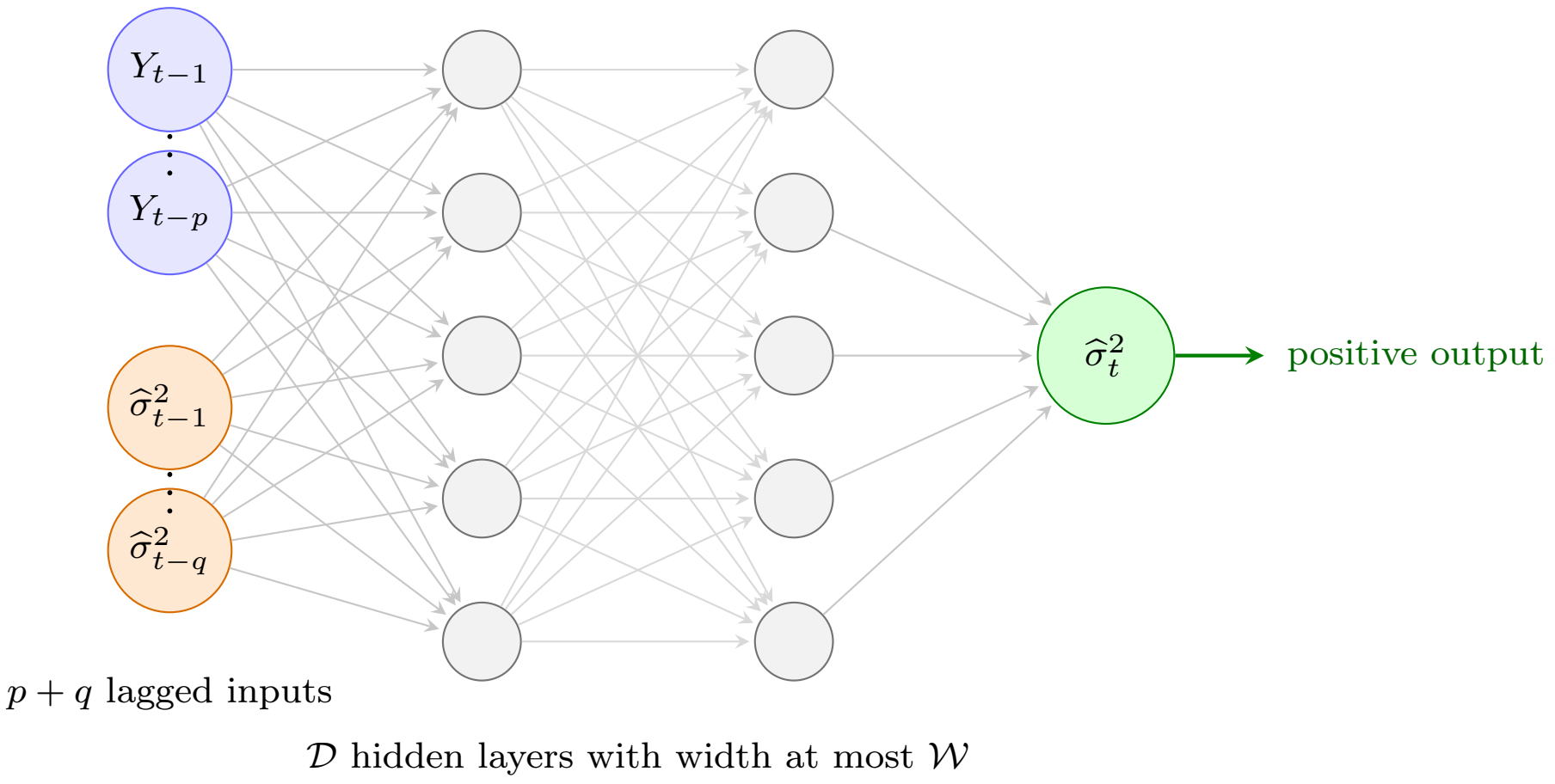
\begin{figure}[t]
\centering
\begin{tikzpicture}[>=stealth, shorten >=1pt,
  ret/.style={circle,draw=blue!60,fill=blue!10,minimum size=0.95cm,inner sep=1pt,font=\scriptsize},
  var/.style={circle,draw=orange!85!black,fill=orange!18,minimum size=0.95cm,inner sep=1pt,font=\scriptsize},
  hid/.style={circle,draw=black!55,fill=black!5,minimum size=0.6cm},
  onode/.style={circle,draw=green!50!black,fill=green!16,minimum size=1.05cm,inner sep=1pt,font=\scriptsize}]
  \node[ret] (I1) at (0,3.3) {$Y_{t-1}$};
  \node[ret] (I2) at (0,2.2) {$Y_{t-p}$};
  \node[var] (I3) at (0,0.7) {$\widehat\sigma^2_{t-1}$};
  \node[var] (I4) at (0,-0.4) {$\widehat\sigma^2_{t-q}$};
  \node at (0,2.75) {$\vdots$};
  \node at (0,0.15) {$\vdots$};
  \foreach \i/\y in {1/3.3,2/2.2,3/1.1,4/0.0,5/-1.1}
    \node[hid] (A\i) at (2.4,\y) {};
  \foreach \i/\y in {1/3.3,2/2.2,3/1.1,4/0.0,5/-1.1}
    \node[hid] (B\i) at (4.8,\y) {};
  \node[onode] (O) at (7.2,1.1) {$\widehat\sigma^2_t$};
  \foreach \a in {I1,I2,I3,I4}
    \foreach \b in {1,2,3,4,5}
      \draw[->,black!22] (\a) -- (A\b);
  \foreach \a in {1,2,3,4,5}
    \foreach \b in {1,2,3,4,5}
      \draw[->,black!15] (A\a) -- (B\b);
  \foreach \a in {1,2,3,4,5}
    \draw[->,black!22] (B\a) -- (O);
  \draw[->,green!50!black,thick] (O) -- ++(1.25,0)
    node[right,font=\scriptsize,green!40!black]
    {positive output};
  \node[font=\scriptsize,align=center] at (0,-1.5) {$p+q$ lagged inputs};
  \node[font=\scriptsize,align=center] at (3.6,-2.0)
    {$\dnet$ hidden layers with width at most $\wnet$};
\end{tikzpicture}
\caption{The neural network used at each iteration.  The inputs are the $p$
return lags and the $q$ variance estimates from the preceding iteration.  The
$\dnet$ hidden layers have maximum width $\wnet$ and map the inputs to a
positive conditional-variance estimate.}
\label{fig:dnn}
\end{figure}

Our iterative algorithm allows a general nonparametric
function class $\Fclass$.  For our neural estimator, we take $\Fclass$ to be
a class of multilayer perceptrons (MLPs), a commonly used class of feedforward
neural networks.
Specifically, for $x\in\R^{p+q}$, an MLP with $\dnet$ hidden
layers has the form
\begin{equation}
 f(x)
 =\psi\circ\mathcal L_{\dnet+1}\circ\phi\circ\mathcal L_{\dnet}
 \circ\cdots\circ\phi\circ\mathcal L_1(x),
 \qquad \mathcal L_\ell(u)=A_\ell u+b_\ell,
 \label{eq:apt-mlp}
\end{equation}
where $A_\ell$ and $b_\ell$ are the weight matrix and bias vector,
respectively, in layer $\ell$.  Here we use the rectified linear unit (ReLU)
activation, $\phi(u)=\max\{u,0\}$, applied coordinatewise.
The output activation $\psi$ takes nonnegative values and
ensures a nonnegative conditional variance estimate.
We denote the number of units in layer $j$ by $d_j$, where $d_0=p+q$ and
$d_{\dnet+1}=1$.  The network depth is $\dnet$.  Its width is
$\wnet=\max_{1\leq j\leq\dnet}d_j$.  We allow some connections to be set to
zero and let $\snet$ denote the number of remaining trainable weights and
biases.  For a fully connected network,
$\snet=\sum_{\ell=1}^{\dnet+1}(d_{\ell-1}+1)d_\ell$.  We allow the depth
$\dnet$, width $\wnet$, and number of trainable parameters $\snet$ to depend
on $T$.  Thus, the network class may expand with the sample size.  We write
$\Fclass=\Fclass_{\dnet,\wnet,\snet}$ for the network class with these
specifications and use $\Fclass_T$ when $(\dnet,\wnet,\snet)$ depend on $T$.
Figure~\ref{fig:dnn} illustrates the network structure.

\revadd{For the nonparametric GARCH model, we use a network class with two
additional properties.} First, each network maps its inputs to a positive bounded interval
containing the range of the true conditional variance.  Projection or a
bounded output activation can produce such outputs and keep all estimated
variance lags in the same interval.  Second, we require each network to be
Lipschitz in its variance inputs.  This property helps to stabilize recursive
forecasts.  For
some $B'\in(0,1)$, the networks satisfy
$|f(y,s)-f(y,s')|\leq B'\lVert s-s'\rVert_\infty$ for every $y\in\R^p$ and
all variance inputs $s$ and $s'$ in the bounded input domain.  In practice,
weight decay can encourage this property, while explicit norm constraints can
enforce it.  Our numerical implementation remains stable without either.

\FloatBarrier
\section{Statistical Properties}
\label{sec:main-results}\label{sec:theory}

In nonparametric GARCH estimation, a useful estimator should accurately estimate
the in-sample latent variance path, recover the recursion map $\ftrue$, and produce
accurate out-of-sample conditional variance forecasts.  This section
studies these three statistical targets.  Section~\ref{sec:master} establishes
a finite-sample upper bound on the time-averaged in-sample error of the estimated
conditional variance path.  It also gives a matching minimax lower bound for
sufficiently smooth recursion maps.  Under additional
stationarity and dependence conditions,
Section~\ref{sec:out-bound} establishes bounds for out-of-sample forecast
error and recursion-map risk.  In every upper-bound result, we show that our
estimator is robust to lag-order overspecification because
fixed fitted orders $p\geq\pstar$ and $q\geq\qstar$
\revadd{would not change the convergence-rate exponent.}

\subsection{Assumptions}
\label{sec:assumptions-body}

To establish an in-sample estimation-error bound, we impose the following
regularity conditions on the data-generating process.

\begin{assumption}[Innovations and Positive Variance]
\label{ass:innov-up}
The innovations $(\varepsilon_t)$ are sub-Gaussian.  For some $\Bmin>0$, the
recursion map satisfies $\ftrue(x)\geq\Bmin$ for every $x\in\R^{\fdim}$.
\end{assumption}

Sub-Gaussianity is a common light-tail condition in nonparametric regression
\citep{bauer2019deep,schmidt2020nonparametric}.  It controls the innovation
contribution to the squared-return regression error and is satisfied, for
example, by Gaussian or bounded innovations. The lower bound on the
conditional variances is the usual
condition that keeps the conditional variance away from zero.
In linear ARCH and GARCH models with nonnegative coefficients, it corresponds
to requiring a strictly positive intercept
\citep{Engle1982,Bollerslev1986}.

\begin{assumption}[H\"older-Smooth Recursion Map]
\label{ass:hol-up}
The conditional variance function $\ftrue$ lies in the H\"older ball
$\mathcal H^\hol(\R^{\fdim},\Bmax)$ of smoothness $\hol$ and radius $\Bmax$
on the regressor space for some $\hol>0$ and $\Bmax>\Bmin$.  A formal definition
of the H\"older ball appears in the supplementary materials.
\end{assumption}

Assumption~\ref{ass:hol-up} imposes a nonparametric regularity condition.
H\"older smoothness is standard in the
approximation theory of deep ReLU networks
\citep{yarotsky2018optimal,petersen2018optimal,lu2020deep,jiao2023deep}.
The smoothness index affects the network approximation error and determines
the nonparametric convergence rate together with the network complexity and
sample size
\citep{bauer2019deep,schmidt2020nonparametric}.

\begin{assumption}[$\sigma$-block Lipschitz of $\ftrue$]
\label{ass:lip-up}
There exist weights $\omega_1,\ldots,\omega_{\qstar}>0$ with
$\sum_{j=1}^{\qstar}\omega_j=1$ and a constant $\Dcon\in[0,1)$ such that
\[
 |\ftrue(y,s)-\ftrue(y,s')|
 \leq \Dcon\lVert s-s'\rVert_{\omega},\qquad
 \lVert s-s'\rVert_{\omega}
 :=\left\{\sum_{j=1}^{\qstar}\omega_j(s_j-s_j')^2\right\}^{1/2},
\]
for all $y\in\R^{\pstar}$ and $s,s'\in\R^{\qstar}$.
\end{assumption}

Assumption~\ref{ass:lip-up} extends the familiar stability restriction on
lagged variance feedback to a nonparametric GARCH setting. It recovers the
usual stability condition in the linear case. For example, suppose that the
conditional variance function is affine,
$\ftrue(y,s)=c_0+\sum_{i=1}^{\pstar}\alpha_i y_i^2+
\sum_{j=1}^{\qstar}\beta_j s_j$, with intercept $c_0>0$ and variance-lag
coefficients $\beta_1,\ldots,\beta_{\qstar}$. If these coefficients are
positive, take $\omega_j=\beta_j/\sum_{\ell=1}^{\qstar}\beta_\ell$. The
weighted Cauchy--Schwarz inequality then gives
$\Dcon=\sum_{j=1}^{\qstar}\beta_j$. Hence Assumption~\ref{ass:lip-up} holds
whenever $\sum_j\beta_j<1$. Because the usual covariance-stationarity condition
$\sum_i\alpha_i+\sum_j\beta_j<1$ implies this requirement, every
covariance-stationary linear GARCH model satisfies
Assumption~\ref{ass:lip-up} \citep{Bollerslev1986}.

\subsection{In-sample estimation error bounds}
\label{sec:master}\label{sec:rate}

We measure the average error of the variance path estimated by $\widehat f_M$
over the estimation sample by
\begin{equation}
 \Delta_{T,M}^{\rm in}
 =\frac{1}{T-r_0}\sum_{t=r_0+1}^{T}
 \E\{\volhat{t,M}-\volsq_t\}^2,
 \label{eq:apt-insample-risk}
\end{equation}
where $r_0=\max(p,q)$ and the expectation is taken over the randomness of the
sample used to train the networks. We next bound this in-sample error.

\begin{theorem}[Nonasymptotic in-sample bound]
\label{thm:main-upfront}
Suppose Assumptions~\ref{ass:innov-up}--\ref{ass:lip-up} hold and
$B'>\Dcon$.  Apply the
iterated neural estimator with input orders $p\geq\pstar$ and $q\geq\qstar$
that do not depend on $T$.  Initialize the unobserved variances at a measurable value
$\widehat c\in[\Bmin,\Bmax]$.  Choose the network size
$\snet\asymp T^{(\pstar+\qstar)/(\pstar+\qstar+2\hol)}$ and take
$M\asymp\log T$.  Then there are constants $C>0$ and $c\geq0$ such that,
for every sufficiently large $T$,
\[
 \Delta_{T,M}^{\rm in}
 \leq C T^{-2\hol/(\fdim+2\hol)}(\log T)^c.
\]
\end{theorem}

Theorem~\ref{thm:main-upfront} establishes that our neural estimator attains
the nonparametric rate with an appropriately chosen network size after
$O(\log T)$ iterations, as the initialization error decays geometrically under
the contraction condition in Assumption~\ref{ass:lip-up}.  Here,
$T^{-2\hol/(2\hol+\pstar+\qstar)}$ is the classical minimax rate for
$\hol$-H\"older nonparametric regression in intrinsic dimension
$\pstar+\qstar$ \citep{stone1982optimal}.  Moreover, our neural estimator is
robust to lag-order overspecification because it retains this rate for any
fixed fitted orders $p\geq\pstar$ and $q\geq\qstar$.  Robustness permits
conservative choices when the true orders are unknown.

Theorem~\ref{thm:main-upfront} establishes consistency of the in-sample
variance path in mean square. \revadd{The closest earlier analysis is given by
\citet{buhlmann2002algorithm}. It concerns GARCH$(1,1)$ and relies on
conditions on the estimated variance inputs that are difficult to verify.} We overcome this
difficulty with a martingale bound that holds uniformly over deterministic
sequences of candidate network functions.  This uniformity controls both the
generated variance regressors and \revadd{the dependence induced by reusing the
same sample across iterations.}

We next establish a lower bound for estimating the same variance path.  Let
$\Cclass$ be the class of conditional variance functions that satisfy the
positive lower bound in Assumption~\ref{ass:innov-up} and the H\"older condition
in Assumption~\ref{ass:hol-up} with $\hol\geq1$.  We assume that every map in
$\Cclass$ has structural lag orders exactly $(\pstar,\qstar)$ and satisfies
Assumption~\ref{ass:lip-up} with a contraction constant $\Dcon\leq\bar D$ for a
common constant $\bar D\in(0,1)$.  For each
$\ftrue\in\Cclass$, we let $\E_{\ftrue}$ denote expectation under the strictly
stationary solution of Model~\ref{model:struct} with independent standard
Gaussian innovations.  We then obtain the following minimax lower bound for
estimating the conditional variance path over $\Cclass$.

\begin{theorem}[Minimax lower bound]
\label{thm:lower}
Set $r_*=\max(\pstar,\qstar)$.  There is a constant
$c_{\rm LB}>0$ such that, for every sufficiently large $T$,
\[
 \inf_{\widetilde{\boldsymbol v}}
 \sup_{\ftrue\in\Cclass}
 \frac1{T-r_*}\sum_{t=r_*+1}^{T}
 \E_{\ftrue}(\widetilde v_t-\volsq_t)^2
 \geq c_{\rm LB}T^{-2\hol/(2\hol+\pstar+\qstar)}.
\]
Here $\widetilde{\boldsymbol v}
=(\widetilde v_{r_*+1},\ldots,\widetilde v_T)$ ranges over all
$\sigma(Y_1,\ldots,Y_T)$-measurable vectors.
\end{theorem}

Theorem~\ref{thm:lower} provides a matching lower bound for the same
time-averaged variance-path loss studied in
Theorem~\ref{thm:main-upfront}.  Hence, our iterated neural
estimator is minimax rate-optimal over $\Cclass$ up to logarithmic factors.
This conclusion remains valid when the fitted lag orders are overspecified.

Theorem~\ref{thm:lower} applies to recursion maps with smoothness
$\hol\geq1$.  This condition allows the lower-bound construction to preserve
contraction in the variance lags.  For $0<\hol<1$, the construction may
violate contraction, and \revadd{a different argument would be needed.}
Related rate results remain limited to first-order recursive models.
\citet{meister2016statistical} study a monotone nonparametric GARCH$(1,1)$
recursion with $\hol>2$.  \citet{wechsung2022consistency} establish
consistency and a convergence rate for a nonparametric INGARCH$(1,1)$ model
with latent intensities.  Neither result provides a matching minimax theory
for the return-based recursion when $0<\hol<1$.

\subsection{Recursion-map recovery and out-of-sample forecasting}
\label{sec:out-bound}

The in-sample result controls the estimated conditional variance path at the
observed dates.  However, it does not establish that the estimated
conditional variance function $\widehat f_M$ is accurate at a new regressor
vector.  We next study the function-estimation error of $\widehat f_M$ and the
conditional variance forecasts based on it.  For this purpose, we impose
additional conditions relating the estimation sample to future observations.

\begin{assumption}[Stationarity and geometric $\beta$-mixing]
\label{ass:dep-up}
The process $(Y_t)_{t\in\Z}$ is strictly stationary and its $\beta$-mixing
coefficients decay geometrically.
\end{assumption}

For conciseness, we impose stationarity and geometric $\beta$-mixing directly in
Assumption~\ref{ass:dep-up}.  More primitive conditions on the volatility
equation and innovation distribution can also imply these properties
\citep{carrasco2002mixing,francq2006mixing,saidi2006stationarity,meitzsaikkonen2008}.

Stationarity and $\beta$-mixing play different roles in our analysis.
For the forecast analysis, stationarity ensures that the historical and future
regressor vectors have the same distribution.  Geometric $\beta$-mixing permits
distant blocks to be coupled
with independent copies.  This construction supports the sample-to-population
comparison and the forecast bound
\citep{berbee1979random,yu1994rates}.  These conditions are sufficient for our
argument but may not be necessary for out-of-sample prediction in general.

We next bound the estimation error of $\widehat f_M$ relative to $\ftrue$ at
a regressor vector constructed from the true conditional variances.  For this
purpose, let
$\{(Y_t',(\sigma_t')^2):t\in\mathbb Z\}$ be a stationary copy of the return
and conditional-variance process, independent of
the estimation sample, and define
\[
 X_{0,*}'=
 \bigl(Y_{-1}',\ldots,Y_{-p}',
 (\sigma_{-1}')^2,\ldots,(\sigma_{-q}')^2\bigr).
\]
If $p>\pstar$ or $q>\qstar$, $\ftrue(X_{0,*}')$ is understood to use only
the first $\pstar$ return coordinates and the first $\qstar$ variance
coordinates of $X_{0,*}'$.

\begin{corollary}[Recursion-map recovery]
\label{cor:func-recovery}
Suppose that the conditions of Theorem~\ref{thm:main-upfront} hold and
Assumption~\ref{ass:dep-up} is satisfied.  For our iterated neural estimator
$\widehat f_M$ with lag orders $p\geq\pstar$ and $q\geq\qstar$ that do not
depend on $T$, we have
\[
 \E\{\widehat f_M(X_{0,*}')-\ftrue(X_{0,*}')\}^2
 \leq C T^{-2\hol/(\fdim+2\hol)}(\log T)^C.
\]
\end{corollary}

The rate matches the classical minimax rate for $\hol$-H\"older nonparametric
regression in dimension $\pstar+\qstar$ \citep{stone1982optimal}.  It also
remains unchanged under lag-order overspecification.

We next study the forecast errors obtained by applying the estimated
conditional variance function $\widehat f_M$ recursively to future
observations.  We use $N\geq0$ post-estimation returns to update the lagged
variance estimates.  We then define the average error of the next $H\geq1$
sequential one-step forecasts as
\begin{equation}
 \Delta_{T,M}^{\rm out}(N,H)
 =\frac1H\sum_{h=1}^{H}
 \E\{\widehat\sigma^2_{T+N+h\mid T}-\volsq_{T+N+h}\}^2,
 \qquad N\geq0,\quad H\geq1.
 \label{eq:apt-oos-risk}
\end{equation}
In particular, $\Delta_{T,M}^{\rm out}(N,1)$ is the single-date risk after
$N$ updates, whereas $\Delta_{T,M}^{\rm out}(0,H)$ averages the first $H$
forecasts after the estimation sample.

\begin{theorem}[Out-of-sample forecasting rates]
\label{thm:oos-upfront}
Suppose that the conditions of Theorem~\ref{thm:main-upfront} hold and
Assumption~\ref{ass:dep-up} is satisfied.  For the forecasts generated by our
iterated neural estimator with lag orders $p\geq\pstar$ and $q\geq\qstar$
that do not depend on $T$, there is a constant
$\lambda_{\mathrm F}\in(0,1)$ such that, for all sufficiently large $T$ and
every $N\geq0$ and $H\geq1$,
\begin{equation}
 \Delta_{T,M}^{\rm out}(N,H)
 \leq C T^{-2\hol/(\fdim+2\hol)}(\log T)^C
 +\frac{C}{H}\sum_{h=1}^H\lambda_{\mathrm F}^{N+h}.
 \label{eq:apt-unified-rate}
\end{equation}
\end{theorem}

Theorem~\ref{thm:oos-upfront} bounds the out-of-sample forecast risk by two
terms.  The first is the mean-squared estimation error of $\widehat f_M$ for
the conditional variance function $\ftrue$.  The second accounts for the
initial variance estimates used in forecasting and for dependence between the
estimation and forecast samples.  For the $h$th forecast after $N$ updates,
this term decreases as $\lambda_{\mathrm F}^{N+h}$.  Here, the decay factor
$\lambda_{\mathrm F}$ depends on the contraction coefficients $\Dcon$ and
$B'$, the variance-lag orders, and the geometric decay rate of $\beta$-mixing,
but not on $T$, $N$, or $H$.  Averaging over $h=1,\ldots,H$ gives the second
term in
\eqref{eq:apt-unified-rate}.

For a single-date forecast ($H=1$), the second term in
\eqref{eq:apt-unified-rate} is $C\lambda_{\mathrm F}^{N+1}$ and is \revadd{no
bigger than} the estimation term once $N$ is a sufficiently large multiple of
$\log T$.  For forecasts beginning immediately
after the estimation sample ($N=0$), averaging over $H$ dates bounds the second
term by $C/H$.  The bound therefore does not give the nonparametric rate for
the first forecast alone.  It yields that rate, up to logarithmic factors,
when $H\gtrsim T^{2\hol/(2\hol+\pstar+\qstar)}$.  As $N\to\infty$, the
initialization and dependence term vanishes geometrically, leaving the
recursion-map rate in Corollary~\ref{cor:func-recovery} for an independent
stationary regressor.  For a finite $N$, increasing $H$ reduces this term's
contribution to the average forecast risk at rate $H^{-1}$.

Lastly, we consider the Gaussian quasi-likelihood (QLIKE) loss.  For a positive
variance forecast $v$ and a nonnegative variance proxy $z$, the QLIKE loss is
defined as $z/v+\log v$ \citep{Patton2011}.  In our model, we use $z=Y_t^2$,
which is conditionally unbiased for $\volsq_t$ because
$\E(Y_t^2\mid\filt{t-1})=\volsq_t$.  The following corollary shows that the
unconditional expectation of the conditional expected excess QLIKE loss
$\volsq_t/v-\log(\volsq_t/v)-1$ is equivalent, up to constants, to the
one-step mean-squared forecast error.

\begin{corollary}[Expected one-step excess QLIKE loss]
\label{cor:qlike-upfront}
Under the conditions of Theorem~\ref{thm:oos-upfront} for fitted orders
$p\geq\pstar$ and $q\geq\qstar$ that do not depend on $T$, for every
$N\geq0$ there are constants
$0<c<C<\infty$, depending only on
$\Bmin$ and $\Bmax$, such that
\[
 c\Delta_{T,M}^{\rm out}(N,1)
 \leq\E\left\{
 \frac{\volsq_{T+N+1}}{\widehat\sigma^2_{T+N+1\mid T}}
 -\log\left(\frac{\volsq_{T+N+1}}
 {\widehat\sigma^2_{T+N+1\mid T}}\right)-1
 \right\}
 \leq C\Delta_{T,M}^{\rm out}(N,1).
\]
Thus, setting $H=1$ in Theorem~\ref{thm:oos-upfront} also bounds the expected
excess QLIKE loss for a one-step forecast.
\end{corollary}

\section{Weighted extension}
\label{sec:weighted-extension}

The estimator in Section~\ref{sec:algo} assigns equal weight to all
observations for which the required lags are available.  This choice is
natural when the conditional variance dynamics are stable.  If the dynamics
evolve, however, older observations may be less informative about the current
conditional variance function.  We therefore extend our estimator by assigning
greater weight to recent observations.

Specifically, let $(a_{T,t}:r_0<t\leq T)$ be deterministic nonnegative weights
with $\sum_{t=r_0+1}^T a_{T,t}=1$.  Starting from the initial variance path in
Section~\ref{sec:algo}, at each iteration $m$ we estimate the network by
weighted least squares,
\begin{equation}
 \widehat f_{m,a}\in\argmin_{f\in\Fclass_T}
 \sum_{t=r_0+1}^{T}a_{T,t}
 \left[
 Y_t^2-f\{Y_{t-1:t-p},\widehat\sigma^2_{t-1:t-q,m-1,a}\}
 \right]^2,
 \label{eq:weighted-erm-main}
\end{equation}
and then update each conditional variance estimate by evaluating
$\widehat f_{m,a}$ at the corresponding return lags and the variance estimates
from iteration $m-1$.
After $M$ iterations, we retain $\widehat f_{M,a}$ and
$(\widehat\sigma^2_{t,M,a})$ as the weighted estimates of the conditional
variance function and variance path.  Equal weights,
$a_{T,t}=(T-r_0)^{-1}$, exactly recover the proposed estimator in
Section~\ref{sec:algo}.

We consider three weighting rules that place more emphasis on
recent observations.  A mild extension of equal weighting assigns linearly
increasing weights, $a_{T,t}\propto t-r_0$, to all available observations.
Exponential weighting uses
$a_{T,t}=\lambda_T^{T-t}/\sum_{s=r_0+1}^{T}\lambda_T^{T-s}$, where
$\lambda_T=2^{-1/h_T}$ and $1\leq h_T\leq T-r_0$.  Moving $h_T$ periods
further into the past halves the weight.  The third rule uses the most recent
$1\leq w_T\leq T-r_0$ observations, each with weight $1/w_T$.

For these deterministic weighting rules, let $s_T$ denote the effective
estimation length. It equals $T-r_0$ for equal and linear weights, is of order
$h_T$ for exponential weights, and equals $w_T$ for window weights. Adapting
the proof of Theorem~\ref{thm:main-upfront} gives an in-sample rate of
$s_T^{-2\hol/(\pstar+\qstar+2\hol)}$ up to logarithmic factors. We can also establish a convergence rate when the recursion map
changes smoothly over time. Section~\ref{sec:sim-time-varying} examines these
weighted extensions.

\section{Simulation studies}
\label{sec_sim}

We compare six estimators under parametric and nonlinear
stationary volatility dynamics and nine under gradual temporal changes. We
use \textbf{MLE}, \textbf{EGARCH}, and \textbf{GJR} as
Gaussian quasi-MLE benchmarks for GARCH, exponential GARCH, and
Glosten--Jagannathan--Runkle GARCH models
\citep{Bollerslev1986,Nelson1991,GlostenJagannathanRunkle1993}. We also
compute the iterated \textbf{Kernel} estimator of
\citet{buhlmann2002algorithm} as a nonparametric benchmark. For the neural
methods, we implement our estimator with an MLP (\textbf{NN-MLP}) and
use a long short-term memory network (\textbf{LSTM}) as a direct
sequence-forecasting benchmark
\citep{hochreiter1997long,kim2018forecasting}. Unlike NN-MLP, LSTM does not
estimate the latent recursion map. In the time-varying
experiment, we compare \textbf{MLE}, \textbf{EGARCH}, \textbf{GJR},
\textbf{Kernel}, \textbf{NN-MLP}, \textbf{LSTM}, and the three weighted
NN-MLP extensions.
\ref{app:simulation-implementation}
gives the implementation details.

For each stationary data-generating process, we discard 2,000 initial
observations and use nested estimation windows of length $T$ with a common
endpoint. We use the next $N=100$ observations to update the variance state
and the following $H=1900$ for evaluation, as in Section~\ref{sec:out-bound}.
We set $T=8000$ unless stated otherwise.
Our main forecast measures are the sample counterparts of the out-of-sample
variance-path risk $\Delta_{T,M}^{\rm out}(N,H)$ in
\eqref{eq:apt-oos-risk} and the latent excess QLIKE loss in
Corollary~\ref{cor:qlike-upfront}.

\subsection{Known lag orders}
\label{sec:sim-known-orders}

We first evaluate these methods under eight volatility designs
with known lag orders.
We set the fitted orders $(p,q)=(p^*,q^*)$ for \textbf{MLE}, \textbf{EGARCH},
\textbf{GJR}, \textbf{Kernel}, and \textbf{NN-MLP}, and use a fixed 20-period
lookback for \textbf{LSTM}. Designs
A--C are order-$(2,2)$ parametric data-generating processes:
{\small
\begin{align*}
 \textbf{A}:\quad \sigma_t^2={}&0.08+0.12Y_{t-1}^2+0.08Y_{t-2}^2
 +0.50\sigma_{t-1}^2+0.18\sigma_{t-2}^2,\\
 \textbf{B}:\quad \sigma_t^2={}&0.10+0.05Y_{t-1}^2+0.03Y_{t-2}^2
 +0.10\mathbf 1_{\{Y_{t-1}<0\}}Y_{t-1}^2\\
 &+0.06\mathbf 1_{\{Y_{t-2}<0\}}Y_{t-2}^2
 +0.45\sigma_{t-1}^2+0.22\sigma_{t-2}^2,\\
 \textbf{C}:\quad \sigma_t^2={}&\exp\Bigl\{-0.18
 +0.22\{|z_{t-1}|-\sqrt{2/\pi}\}
 +0.10\{|z_{t-2}|-\sqrt{2/\pi}\}\\
 &-0.12z_{t-1}-0.06z_{t-2}
 +0.45\log\sigma_{t-1}^2+0.20\log\sigma_{t-2}^2\Bigr\},
\end{align*}}where $z_{t-j}=Y_{t-j}/\sqrt{\sigma_{t-j}^2}$. Thus
\textbf{MLE}, \textbf{GJR}, and \textbf{EGARCH} are correctly specified for A, B,
and C, respectively. \revadd{We also consider five nonlinear recursion-map
designs, D--H, with details given in \ref{app:simulation-dgps}.}
Designs D--H have respective orders
$(3,3)$, $(3,4)$, $(4,3)$, $(4,4)$, and $(5,5)$. They incorporate smooth
threshold asymmetry, interactions among lagged returns, and state-dependent
feedback from lagged variances.

\begin{table}[!t]
\centering
\caption{Out-of-sample performance under stationary dynamics. Panel A reports
variance-path MSE, and Panel B reports latent excess QLIKE. Entries are medians
over 50 replications, with Monte Carlo standard deviations in parentheses.
Values are scaled by 100.}
\label{tab:sim-known-path-new}
\label{tab:sim-known-qlike-new}
\footnotesize
\setlength{\tabcolsep}{2.5pt}
\renewcommand{\arraystretch}{1.08}
\begin{tabular*}{\textwidth}{@{\extracolsep{\fill}}lcccccccc@{}}
\toprule
\textit{Panel A.} & \multicolumn{3}{c}{Parametric designs} & \multicolumn{5}{c}{Nonlinear designs} \\
\cmidrule(lr){2-4}\cmidrule(lr){5-9}
& A(2,2) & B(2,2) & C(2,2) & D(3,3) & E(3,4) & F(4,3) & G(4,4) & H(5,5) \\
\midrule
MLE & \shortstack{\textbf{0.08}\\[-1pt] \textbf{(0.12)}} & \shortstack{0.93\\[-1pt] (0.21)} & \shortstack{1.77\\[-1pt] (0.22)} & \shortstack{8.04\\[-1pt] (0.71)} & \shortstack{8.93\\[-1pt] (2.01)} & \shortstack{12.09\\[-1pt] (0.99)} & \shortstack{10.36\\[-1pt] (1.09)} & \shortstack{3.77\\[-1pt] (0.25)} \\
EGARCH & \shortstack{0.72\\[-1pt] (1.55)} & \shortstack{0.49\\[-1pt] (0.32)} & \shortstack{\textbf{0.08}\\[-1pt] \textbf{(0.08)}} & \shortstack{3.28\\[-1pt] (0.53)} & \shortstack{4.86\\[-1pt] (1.58)} & \shortstack{10.84\\[-1pt] (0.81)} & \shortstack{4.65\\[-1pt] (1.42)} & \shortstack{3.47\\[-1pt] (0.21)} \\
GJR & \shortstack{0.15\\[-1pt] (0.16)} & \shortstack{\textbf{0.09}\\[-1pt] \textbf{(0.06)}} & \shortstack{0.51\\[-1pt] (0.46)} & \shortstack{6.89\\[-1pt] (1.25)} & \shortstack{9.00\\[-1pt] (2.26)} & \shortstack{12.53\\[-1pt] (1.00)} & \shortstack{14.51\\[-1pt] (5.33)} & \shortstack{3.88\\[-1pt] (0.26)} \\
Kernel & \shortstack{4.35\\[-1pt] (5.12)} & \shortstack{2.65\\[-1pt] (1.53)} & \shortstack{1.23\\[-1pt] (0.44)} & \shortstack{3.80\\[-1pt] (0.96)} & \shortstack{3.59\\[-1pt] (0.97)} & \shortstack{7.59\\[-1pt] (1.22)} & \shortstack{4.92\\[-1pt] (0.91)} & \shortstack{2.61\\[-1pt] (0.47)} \\
NN-MLP & \shortstack{1.62\\[-1pt] (2.15)} & \shortstack{0.86\\[-1pt] (1.27)} & \shortstack{0.57\\[-1pt] (0.22)} & \shortstack{\textbf{2.92}\\[-1pt] \textbf{(0.57)}} & \shortstack{\textbf{2.14}\\[-1pt] \textbf{(0.40)}} & \shortstack{\textbf{5.13}\\[-1pt] \textbf{(0.76)}} & \shortstack{\textbf{2.31}\\[-1pt] \textbf{(0.54)}} & \shortstack{\textbf{2.21}\\[-1pt] \textbf{(0.29)}} \\
LSTM & \shortstack{1.07\\[-1pt] (2.99)} & \shortstack{0.71\\[-1pt] (0.72)} & \shortstack{0.41\\[-1pt] (0.28)} & \shortstack{3.66\\[-1pt] (1.38)} & \shortstack{3.26\\[-1pt] (1.02)} & \shortstack{11.64\\[-1pt] (1.63)} & \shortstack{5.69\\[-1pt] (4.10)} & \shortstack{3.60\\[-1pt] (0.66)} \\
\midrule
\textit{Panel B.} & A(2,2) & B(2,2) & C(2,2) & D(3,3) & E(3,4) & F(4,3) & G(4,4) & H(5,5) \\
\midrule
MLE & \shortstack{\textbf{0.06}\\[-1pt] \textbf{(0.04)}} & \shortstack{0.75\\[-1pt] (0.07)} & \shortstack{1.58\\[-1pt] (0.10)} & \shortstack{3.87\\[-1pt] (0.17)} & \shortstack{3.86\\[-1pt] (0.22)} & \shortstack{6.10\\[-1pt] (0.26)} & \shortstack{9.02\\[-1pt] (0.49)} & \shortstack{3.19\\[-1pt] (0.15)} \\
EGARCH & \shortstack{0.44\\[-1pt] (0.10)} & \shortstack{0.36\\[-1pt] (0.07)} & \shortstack{\textbf{0.08}\\[-1pt] \textbf{(0.05)}} & \shortstack{\textbf{1.38}\\[-1pt] \textbf{(0.08)}} & \shortstack{1.91\\[-1pt] (0.13)} & \shortstack{5.43\\[-1pt] (0.22)} & \shortstack{\textbf{1.79}\\[-1pt] \textbf{(0.14)}} & \shortstack{2.90\\[-1pt] (0.13)} \\
GJR & \shortstack{0.09\\[-1pt] (0.05)} & \shortstack{\textbf{0.07}\\[-1pt] \textbf{(0.04)}} & \shortstack{0.33\\[-1pt] (0.05)} & \shortstack{2.66\\[-1pt] (0.15)} & \shortstack{3.39\\[-1pt] (0.23)} & \shortstack{6.20\\[-1pt] (0.25)} & \shortstack{3.86\\[-1pt] (0.32)} & \shortstack{3.28\\[-1pt] (0.15)} \\
Kernel & \shortstack{2.92\\[-1pt] (1.71)} & \shortstack{2.23\\[-1pt] (0.81)} & \shortstack{1.15\\[-1pt] (0.55)} & \shortstack{1.92\\[-1pt] (0.72)} & \shortstack{2.55\\[-1pt] (0.81)} & \shortstack{3.90\\[-1pt] (0.83)} & \shortstack{4.80\\[-1pt] (0.83)} & \shortstack{2.28\\[-1pt] (0.77)} \\
NN-MLP & \shortstack{0.81\\[-1pt] (0.32)} & \shortstack{0.74\\[-1pt] (0.40)} & \shortstack{0.59\\[-1pt] (0.19)} & \shortstack{1.53\\[-1pt] (0.29)} & \shortstack{\textbf{1.54}\\[-1pt] \textbf{(0.28)}} & \shortstack{\textbf{2.97}\\[-1pt] \textbf{(0.51)}} & \shortstack{2.54\\[-1pt] (0.91)} & \shortstack{\textbf{1.95}\\[-1pt] \textbf{(0.62)}} \\
LSTM & \shortstack{0.63\\[-1pt] (0.60)} & \shortstack{0.38\\[-1pt] (0.35)} & \shortstack{0.40\\[-1pt] (0.18)} & \shortstack{1.97\\[-1pt] (0.45)} & \shortstack{3.11\\[-1pt] (0.68)} & \shortstack{6.13\\[-1pt] (0.66)} & \shortstack{5.52\\[-1pt] (1.01)} & \shortstack{3.12\\[-1pt] (0.53)} \\
\bottomrule
\end{tabular*}
\end{table}
The correctly specified \textbf{MLE}, \textbf{GJR}, and
\textbf{EGARCH} naturally perform best in A--C. The
\textbf{NN-MLP} estimator has the smallest median variance-path MSE in all five
nonlinear designs and the smallest median latent excess QLIKE in E, F, and H.
Tables~\ref{tab:sim-known-insample} and \ref{tab:sim-sample-size-oos} in
\ref{app:simulation-extra} report in-sample and out-of-sample variance-path
MSEs for $T=2000,4000,8000$.

\subsection{Unknown lag orders}
\label{sec:sim-order-sensitivity}

\begin{figure}[!t]
\centering
\includegraphics[width=\textwidth]{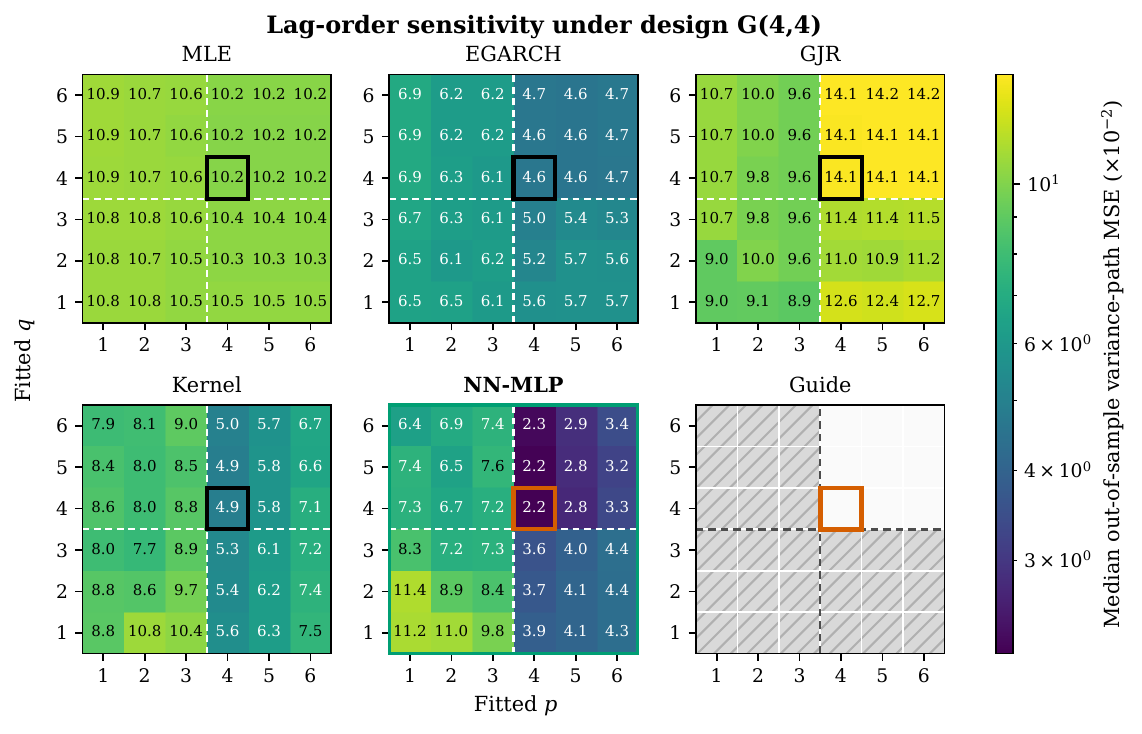}
\caption{Lag-order sensitivity under design G$(4,4)$. Cell entries are median
out-of-sample variance-path MSEs in units of $10^{-2}$ over 20 replications.
Dashed lines identify fitted orders below the structural values, and boxes mark
the structural order. The shaded area in the guide marks fitted orders below at
least one structural value.}
\label{fig:lag-order-g}
\end{figure}

True lag orders are usually unknown. Theorem~\ref{thm:main-upfront} shows that
overspecification preserves the convergence rate. We examine design G$(4,4)$
for $(p,q)\in\{1,\ldots,6\}^2$, comparing MLE, EGARCH, GJR, Kernel, and
NN-MLP over 20 paired replications. LSTM has no GARCH lag-order pair.
Figure~\ref{fig:lag-order-g} shows that NN-MLP is much more sensitive to
omitted than additional lags.

\subsection{Time-varying dynamics}
\label{sec:sim-time-varying}

We next examine \textbf{NN-MLP}, \textbf{LSTM}, and three weighted extensions
(\textbf{NN-Linear} with linear weights, \textbf{NN-EW} with
exponentially decaying weights, and \textbf{NN-Window} with equal recent-window
weights) under three gradual temporal-change designs.
Specifically, we use two target functions $f_-$ and $f_+$ from
a nonlinear GARCH$(2,2)$ family to generate three dynamic designs. The
functions emphasize negative- and positive-return effects, respectively. See
Appendix A.1 for the formulas. In the scale-change design, we generate the
variance sequence from a time-varying multiple of $f_-$. The multiplier
increases smoothly from 0.65 to 1.45 over the transition segment. In the
function-change design, the map changes smoothly from $f_-$ to $f_+$. The
joint-change design combines these two changes.
For each design, we discard 1,000 initial observations and
retain the next 4,000 observations for estimation. It has
1,200 initial observations, 2,400 transition observations, and 400 final
observations. The terminal map continues over the
2,000-observation forecast segment, corresponding to $(N,H)=(0,2{,}000)$ in
Section~\ref{sec:out-bound}.
\begin{table}[!t]
\centering
\caption{Out-of-sample performance under temporal change. The
two blocks report variance-path MSE and latent excess QLIKE. Entries are
medians over 50 replications, with Monte Carlo standard deviations in
parentheses. Values are scaled by 100.}
\label{tab:sim-drift-new}
\scriptsize
\setlength{\tabcolsep}{2pt}
\renewcommand{\arraystretch}{1.08}
\begin{tabular*}{\textwidth}{@{\extracolsep{\fill}}llccccccccc@{}}
\toprule
Metric & Design & MLE & EGARCH & GJR & Kernel & \shortstack{NN-\\MLP}
& \shortstack{NN-\\Linear} & \shortstack{NN-\\EW}
& \shortstack{NN-\\Window} & LSTM \\
\midrule
\multirow{3}{*}{\shortstack[l]{Future\\path\\MSE}}
& Scale
& \shortstack{18.01\\[-1pt] (5.17)}
& \shortstack{11.32\\[-1pt] (2.18)}
& \shortstack{18.11\\[-1pt] (5.80)}
& \shortstack{16.83\\[-1pt] (2.70)}
& \shortstack{6.50\\[-1pt] (1.69)}
& \shortstack{5.54\\[-1pt] (1.97)}
& \shortstack{5.38\\[-1pt] (2.18)}
& \shortstack{\textbf{4.79}\\[-1pt] \textbf{(2.11)}}
& \shortstack{5.03\\[-1pt] (2.09)} \\
& Function
& \shortstack{2.58\\[-1pt] (0.68)}
& \shortstack{1.79\\[-1pt] (0.40)}
& \shortstack{3.01\\[-1pt] (0.77)}
& \shortstack{2.13\\[-1pt] (0.34)}
& \shortstack{1.43\\[-1pt] (0.35)}
& \shortstack{1.02\\[-1pt] (0.36)}
& \shortstack{\textbf{0.97}\\[-1pt] \textbf{(0.40)}}
& \shortstack{1.28\\[-1pt] (0.53)}
& \shortstack{1.27\\[-1pt] (0.34)} \\
& Joint
& \shortstack{16.69\\[-1pt] (5.06)}
& \shortstack{21.05\\[-1pt] (6.01)}
& \shortstack{17.51\\[-1pt] (5.12)}
& \shortstack{23.85\\[-1pt] (6.41)}
& \shortstack{9.89\\[-1pt] (2.69)}
& \shortstack{9.16\\[-1pt] (2.51)}
& \shortstack{8.88\\[-1pt] (2.34)}
& \shortstack{7.04\\[-1pt] (2.51)}
& \shortstack{\textbf{6.77}\\[-1pt] \textbf{(2.65)}} \\
\addlinespace[2pt]
\multirow{3}{*}{\shortstack[l]{Latent\\
excess\\QLIKE}}
& Scale
& \shortstack{8.89\\[-1pt] (1.35)}
& \shortstack{4.59\\[-1pt] (0.84)}
& \shortstack{8.69\\[-1pt] (1.41)}
& \shortstack{35.98\\[-1pt] (7.47)}
& \shortstack{11.27\\[-1pt] (3.13)}
& \shortstack{7.56\\[-1pt] (2.53)}
& \shortstack{5.60\\[-1pt] (2.05)}
& \shortstack{\textbf{3.11}\\[-1pt]
\textbf{(1.38)}}
& \shortstack{5.44\\[-1pt] (3.61)} \\
& Function
& \shortstack{4.90\\[-1pt] (0.67)}
& \shortstack{4.42\\[-1pt] (0.77)}
& \shortstack{5.36\\[-1pt] (0.79)}
& \shortstack{8.40\\[-1pt] (1.79)}
& \shortstack{4.99\\[-1pt] (1.12)}
& \shortstack{2.83\\[-1pt] (1.51)}
& \shortstack{\textbf{2.29}\\[-1pt]
\textbf{(0.82)}}
& \shortstack{2.52\\[-1pt] (4.68)}
& \shortstack{3.34\\[-1pt] (0.99)} \\
& Joint
& \shortstack{5.16\\[-1pt] (1.20)}
& \shortstack{4.55\\[-1pt] (0.52)}
& \shortstack{5.48\\[-1pt] (1.23)}
& \shortstack{24.24\\[-1pt] (7.72)}
& \shortstack{6.61\\[-1pt] (2.38)}
& \shortstack{5.65\\[-1pt] (1.89)}
& \shortstack{4.74\\[-1pt] (1.46)}
& \shortstack{\textbf{2.38}\\[-1pt]
\textbf{(1.39)}}
& \shortstack{4.39\\[-1pt] (3.24)} \\
\bottomrule
\end{tabular*}
\end{table}
Overall, Table~\ref{tab:sim-drift-new} shows that at least one
weighted NN-MLP extension improves on NN-MLP under both out-of-sample losses
in each temporal-change design.

\section{Real-data applications}
\label{sec:real-data}

\begin{figure}[ht!]
\centering
\includegraphics[width=.95\textwidth]{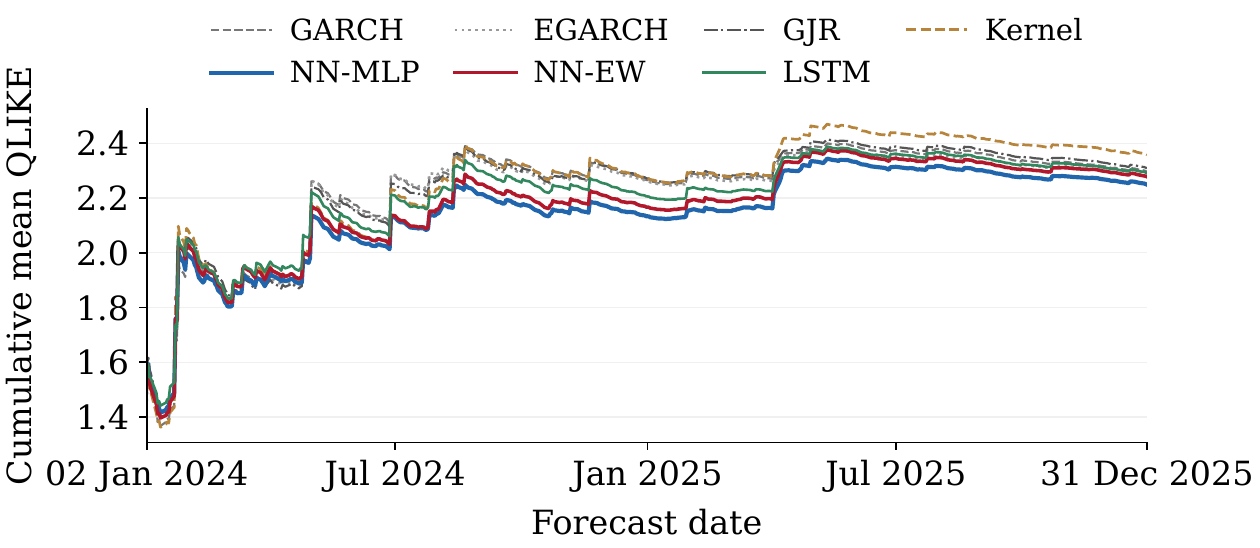}
\caption{Running average test QLIKE for the eight-stock daily panel. Each curve
first averages QLIKE equally across stocks on each date and then averages
through the corresponding forecast date.}
\label{fig:daily-loss-paths}
\end{figure}

\noindent\textbf{Daily volatility forecasts.}
\revadd{To assess the proposed method's performance in daily conditional
variance forecasting, we screen S\&P 100 stocks using the BDS test
\citep{BrockDechertScheinkmanLeBaron1996}. Specifically, we apply the test
to returns standardized by GARCH-type models fitted through 2021 and select
eight eligible stocks with the largest BDS statistics:
3M (MMM), Union Pacific (UNP), International Business Machines (IBM), Nike (NKE),
NextEra Energy (NEE), Target (TGT), Starbucks (SBUX), and Kraft Heinz (KHC).
(See \ref{app:real-reference} for details on stock selection.)}
\revadd{We use past returns and variance estimates to forecast the next-day
conditional variance of percentage log returns. Because the target is
unobserved, we evaluate forecasts against realized variance, calculated as
the sum of squared overnight and five-minute intraday returns. Models are
fitted on 2016--2023 data and evaluated over 2024--2025.}
We report MSE and QLIKE and express returns in percent when computing
variance MSE.
\revadd{Figure~\ref{fig:daily-loss-paths} presents the running average QLIKE across the eight stocks, and
Table~\ref{tab:daily-architecture} reports the test QLIKE and variance MSE for
these eight stocks.}

\begin{table}[ht!]
{
\centering
\caption{Test QLIKE and variance MSE for eight stocks. Each entry reports
QLIKE, with variance MSE in parentheses. Both values are rounded to two decimal
places.}
\label{tab:daily-architecture}
\scriptsize
\setlength{\tabcolsep}{3pt}
\renewcommand{\arraystretch}{1.15}
\resizebox{\textwidth}{!}{%
\begin{tabular}{l*{8}{c}}
\toprule
Method & MMM & UNP & IBM & NKE & NEE & TGT & SBUX & KHC\\
\midrule
GARCH
& \shortstack{2.26\\[-1pt] (57.26)}
& \shortstack{1.64\\[-1pt] (12.91)}
& \shortstack{2.28\\[-1pt] (78.95)}
& \shortstack{2.95\\[-1pt] (597.44)}
& \shortstack{2.03\\[-1pt] (12.94)}
& \shortstack{2.80\\[-1pt] (567.18)}
& \shortstack{2.71\\[-1pt] (358.59)}
& \shortstack{1.71\\[-1pt] (11.38)}\\
EGARCH
& \shortstack{2.22\\[-1pt] (54.06)}
& \shortstack{1.62\\[-1pt] (12.77)}
& \shortstack{\textbf{2.24}\\[-1pt] \textbf{(77.26)}}
& \shortstack{2.94\\[-1pt] (1018.43)}
& \shortstack{2.03\\[-1pt] (12.76)}
& \shortstack{2.82\\[-1pt] (5549.27)}
& \shortstack{2.65\\[-1pt] (341.79)}
& \shortstack{1.72\\[-1pt] (11.49)}\\
GJR
& \shortstack{2.24\\[-1pt] (55.34)}
& \shortstack{\textbf{1.59}\\[-1pt] \textbf{(12.24)}}
& \shortstack{2.28\\[-1pt] (79.78)}
& \shortstack{2.93\\[-1pt] (622.61)}
& \shortstack{2.06\\[-1pt] (13.88)}
& \shortstack{2.85\\[-1pt] (715.46)}
& \shortstack{2.80\\[-1pt] (361.26)}
& \shortstack{1.73\\[-1pt] (12.04)}\\
Kernel
& \shortstack{2.39\\[-1pt] (54.14)}
& \shortstack{1.72\\[-1pt] (14.48)}
& \shortstack{2.34\\[-1pt] (78.62)}
& \shortstack{3.05\\[-1pt] (571.12)}
& \shortstack{2.05\\[-1pt] (13.10)}
& \shortstack{2.86\\[-1pt] (514.95)}
& \shortstack{2.77\\[-1pt] (344.04)}
& \shortstack{1.70\\[-1pt] (11.23)}\\
NN-MLP
& \shortstack{2.27\\[-1pt] (53.80)}
& \shortstack{1.74\\[-1pt] (14.44)}
& \shortstack{2.31\\[-1pt] (78.80)}
& \shortstack{2.78\\[-1pt] (564.27)}
& \shortstack{2.02\\[-1pt] (12.85)}
& \shortstack{\textbf{2.64}\\[-1pt] (509.84)}
& \shortstack{\textbf{2.55}\\[-1pt] \textbf{(341.19)}}
& \shortstack{\textbf{1.69}\\[-1pt] \textbf{(10.90)}}\\
NN-EW
& \shortstack{2.30\\[-1pt] \textbf{(53.28)}}
& \shortstack{1.73\\[-1pt] (14.38)}
& \shortstack{2.48\\[-1pt] (79.67)}
& \shortstack{\textbf{2.72}\\[-1pt] \textbf{(562.52)}}
& \shortstack{\textbf{2.02}\\[-1pt] \textbf{(12.53)}}
& \shortstack{2.69\\[-1pt] \textbf{(509.62)}}
& \shortstack{2.57\\[-1pt] (341.55)}
& \shortstack{1.71\\[-1pt] (11.02)}\\
LSTM
& \shortstack{\textbf{2.19}\\[-1pt] (54.49)}
& \shortstack{1.69\\[-1pt] (13.82)}
& \shortstack{2.38\\[-1pt] (77.76)}
& \shortstack{2.86\\[-1pt] (566.61)}
& \shortstack{2.12\\[-1pt] (13.80)}
& \shortstack{2.70\\[-1pt] (511.02)}
& \shortstack{2.62\\[-1pt] (343.78)}
& \shortstack{1.77\\[-1pt] (12.16)}\\
\bottomrule
\end{tabular}%
}
}
\end{table}
\FloatBarrier
\revadd{Overall, the neural estimators compare favorably with the GARCH-type
benchmarks, with NN-MLP attaining the lowest average QLIKE and NN-EW the
lowest average variance MSE.}

\par\medskip\noindent\textbf{Option-price information.}
\revadd{We also examine whether information from option prices
\citep[see also][]{ChongTodorov2024} improves next-day variance forecasts
for the same eight stocks. We use implied
volatility inferred from option prices
\citep{ChristensenPrabhala1998,BlairPoonTaylor2001} for each target stock and
an auxiliary stock. For each target, we choose the auxiliary from the other
seven stocks based on the highest historical daily-return correlation.}
\revadd{We compare NN-MLP, NN-EW, Kernel, and LSTM trained to predict either
squared returns $Y_t^2$ or realized variance $RV_t$.
We consider three nested information sets. M0 uses past target returns,
their squares for LSTM, and past variance estimates for the other methods.
M1 adds the previous day's target implied volatility and VIX, a market-wide
measure of expected volatility derived from S\&P 500 option prices. M2 also
adds a measure of changes in the auxiliary stock's implied volatility.
See \ref{app:real-reference} for implementation details.}

\begin{table}[ht!]
{
\centering
\caption{Average next-day realized-variance forecast errors for the eight
stocks under three information sets. Entries report mean QLIKE across stocks,
with mean variance MSE in parentheses. All specifications use the same 155
test dates.}
\label{tab:iv-main}
\small
\setlength{\tabcolsep}{3.75pt}
\renewcommand{\arraystretch}{0.88}
\begin{tabular}{@{}llccc@{}}
\toprule
Training response & Method & M0 & M1 & M2\\
\midrule
\multirow{4}{*}{$Y_t^2$} & Kernel & 2.724 (397.67) & 2.556 (395.91) & 2.496 (386.86)\\
 & NN-MLP & 2.761 (394.98) & \textbf{2.467} \textbf{(383.69)} & \textbf{2.472} \textbf{(384.36)}\\
 & NN-EW & 2.812 (396.94) & 2.566 (389.93) & 2.575 (389.38)\\
 & LSTM & \textbf{2.697} \textbf{(394.95)} & 2.682 (394.35) & 2.682 (394.16)\\
\addlinespace[3pt]
\multirow{4}{*}{$RV_t$} & Kernel & 2.611 (393.21) & 2.478 (389.54) & 2.490 (390.45)\\
 & NN-MLP & 2.602 (392.24) & \textbf{2.423} \textbf{(378.37)} & \textbf{2.416} \textbf{(372.00)}\\
 & NN-EW & \textbf{2.582} \textbf{(391.63)} & 2.430 (380.26) & 2.419 (376.80)\\
 & LSTM & 2.625 (393.33) & 2.583 (392.38) & 2.564 (391.90)\\
\bottomrule
\end{tabular}
}
\end{table}
\FloatBarrier

\revadd{Adding option-price information generally lowers both forecast errors.
Using realized variance as the training response lowers both errors for every
method and information set. Under M2 with this response, NN-MLP and NN-EW
outperform Kernel and LSTM on average.}

\par\medskip
\begingroup
\clubpenalty=10000
\noindent\textbf{Five-minute variance forecasts.}
\revadd{We next evaluate the methods on high-frequency data by forecasting
five-minute conditional variances. We focus on MMM and UNP, whose auxiliary
stocks in the preceding application are IBM and MMM, respectively.
See \ref{app:real-reference} for implementation details.}
\revadd{We compute intraday returns from midpoint prices. Because five-minute
conditional variance is unobserved, we evaluate forecasts against realized
variance, calculated as the sum of five squared one-minute returns
\citep{AndersenBollerslev1998}.}
\revadd{We compare two nested information sets. The baseline uses past target
returns, as in M0. The augmented set adds the two preceding realized
variances, the preceding auxiliary-stock return, the preceding count and mean
spacing of midpoint-price changes, and time of day.}
\par\endgroup
\begin{table}[ht!]
{
\centering
\caption{Test losses for five-minute forecasts of MMM and UNP. Entries report
QLIKE, with variance MSE in millions in parentheses. Both stocks have 200
common test days and 72 evaluated intervals per day.}
\label{tab:five-minute-main}
\small
\setlength{\tabcolsep}{2pt}
\renewcommand{\arraystretch}{1.05}
\begin{tabular*}{\textwidth}{@{\extracolsep{\fill}}lcccc@{}}
\toprule
& \multicolumn{2}{c}{MMM} & \multicolumn{2}{c}{UNP}\\
\cmidrule(lr){2-3}\cmidrule(lr){4-5}
Method & Baseline & Augmented & Baseline & Augmented\\
\midrule
Kernel & 6.0178 (0.8866) & 6.0427 (0.8948) & 5.6890 (0.5308) & 5.6560 (0.5157)\\
NN-MLP & 6.0142 (\textbf{0.8526}) & 5.9850 (\textbf{0.8245}) & \textbf{5.6249} (\textbf{0.4992}) & 5.6081 (0.4928)\\
NN-EW & 6.0661 (0.8711) & \textbf{5.9682} (0.8307) & 5.7167 (0.5273) & \textbf{5.6059} (\textbf{0.4919})\\
LSTM & \textbf{6.0100} (0.8670) & 6.0189 (0.8430) & 5.6801 (0.5169) & 5.6415 (0.4966)\\
\bottomrule
\end{tabular*}
}
\end{table}

\FloatBarrier
\revadd{For both stocks, added predictors lower both losses for NN-MLP and
NN-EW, which then outperform Kernel and LSTM on both losses.}

\section{Conclusion}
\label{sec:conclusion}

We propose an iterated neural estimator for the recursion map and latent
conditional variances in nonparametric GARCH models.  We also develop a
weighted extension that emphasizes recent observations.  Without stationarity
or mixing, the in-sample variance-path error attains the standard nonparametric
rate up to logarithmic factors.  A matching lower bound for $\beta\geq1$
establishes minimax optimality.  Under strict stationarity and geometric
$\beta$-mixing, the recursion-map error attains the same polynomial rate.  The
forecast bound adds a geometrically decreasing term due to forecast
initialization and temporal dependence and controls
expected excess QLIKE loss.  These rates depend on the intrinsic lag orders and
are unchanged when the fitted lag orders are larger but do not grow with $T$.
Simulations show that the neural estimators perform well under nonlinear
dynamics and that weighting recent observations improves forecasts under
temporal change.
The real-data applications provide evidence on the
practical performance of the neural estimators. They perform favorably across
the selected stocks. Future work may extend the theory to heavy-tailed innovations and long-memory
dependence. It may also incorporate realized measures through a measurement
equation that accounts for high-frequency measurement noise.

\appendix
\section{Simulation details and additional results}
\label{app:simulation-extra}

\subsection{Data-generating processes}
\label{app:simulation-dgps}

All stationary designs satisfy
$Y_t=\sigma_t\varepsilon_t$ with i.i.d. standard Gaussian $\varepsilon_t$.
Section~\ref{sec:sim-known-orders} defines A--C. For D--H, define
$r_j=Y_{t-j}^2/(0.25+Y_{t-j}^2)$,
$u_j=\sigma_{t-j}^2/(1+\sigma_{t-j}^2)$,
$n_j=\{1+\exp(5Y_{t-j})\}^{-1}$, and $\bar n_j=1-n_j$.
{\small
\begin{align*}
 \textbf{D}(3,3):\quad \sigma_t^2={}&0.05+0.14r_1+0.06r_2+0.20r_3
 +0.38n_1r_3+0.30n_3r_1\\
 &+0.03\sigma_{t-1}^2+0.02\sigma_{t-2}^2
 +(0.35+0.25n_1)\sigma_{t-3}^2,\\[2pt]
 \textbf{E}(3,4):\quad \sigma_t^2={}&0.08+
 \Bigl(0.08+0.92\bigl[1-\exp\{-1.10(\lvert Y_{t-1}\rvert
 +0.80\lvert Y_{t-3}\rvert)\}\bigr]\Bigr)\\
 &\times\Bigl\{0.14+0.30r_1+0.20r_2+0.18r_3+0.28n_1\bar n_2\\
 &\hspace{8mm}+0.03\sigma_{t-1}^2+0.02\sigma_{t-2}^2
 +0.02\sigma_{t-3}^2+(0.38+0.22n_1)\sigma_{t-4}^2\Bigr\}.
\end{align*}
}
For F, define
$\mathcal R_F=n_1\bar n_4+\bar n_1n_4$ and
$\mathcal S_F=n_1n_2n_3+\bar n_1\bar n_2\bar n_3$.
{\small
\begin{align*}
 \textbf{F}(4,3):\quad \sigma_t^2={}&0.05+(0.05+0.55\mathcal R_F)r_4
 +(0.04+0.45\mathcal S_F)r_1\\
 &+0.32r_2r_3+0.20\mathcal R_F\,r_2
 +0.03\sigma_{t-1}^2+0.02\sigma_{t-2}^2\\
 &+(0.34+0.24\mathcal R_F)\sigma_{t-3}^2.
\end{align*}
}
For G, also define
$h_j=\tfrac18[\log\{1+\exp(-8Y_{t-j})\}
-\log\{1+\exp(-8(Y_{t-j}+1))\}]$.
{\small
\begin{align*}
 \textbf{G}(4,4):\quad \sigma_t^2={}&0.18+0.08(h_1+h_2+h_3)+0.60h_4
 +0.42h_1h_4\\
 &+0.30h_2h_3+0.12h_1r_2
 +\frac{0.58+0.14h_1}{1+\exp\{-5(\sigma_{t-4}^2-0.90)\}}.
\end{align*}
}
For H, define
$\mathcal S_H=n_1n_2n_3+\bar n_1\bar n_2\bar n_3$,
$\mathcal A_H=n_1\bar n_2n_3\bar n_4+
\bar n_1n_2\bar n_3n_4$, and
$\mathcal R_H=n_1\bar n_5+\bar n_1n_5$.
{\small
\begin{align*}
 \textbf{H}(5,5):\quad \sigma_t^2={}&0.09+0.20\Bigl\{1-\prod_{j=1}^5(1-r_j)\Bigr\}
 +0.65\mathcal S_H\,r_5+0.55\mathcal A_H\,r_1\\
 &+0.35\mathcal R_H\,r_3+0.28r_2r_4
 +(0.05+0.22\mathcal S_H)u_1+0.04\sum_{j=2}^4u_j\\
 &+(0.06+0.26\mathcal A_H+0.13\mathcal R_H)u_5.
\end{align*}
}
The bounded return terms and variance-input contraction bounds
$0.65$, $0.67$, $0.63$, $0.90$, and $0.84$, respectively, imply bounded
invariant variance intervals for D--H. The maps are Lipschitz on these
intervals.

The experiment in Section~\ref{sec:sim-time-varying} uses
nonlinear GARCH$(2,2)$ recursions. Their return effects are bounded, and their
variance feedback is stable.
For $a>0$, write $r_a(y)=y^2/(a+y^2)$. Evaluated at the true regressor
$\Xtrue{t}$, the two variance functions are
{\small
\begin{align*}
 f_-(\Xtrue{t})={}&0.05+r_{0.50}(Y_{t-1})
 \Bigl[0.24+0.15\{1-\tanh(3Y_{t-1})\}\\
 &\hspace{10mm}+0.16\frac{\sigma_{t-1}^2}{0.50+\sigma_{t-1}^2}\Bigr]
 +0.08r_{1.00}(Y_{t-2})
 +0.30\sigma_{t-1}^2+0.08\sigma_{t-2}^2,\\
 f_+(\Xtrue{t})={}&0.10
 +r_{0.75}(Y_{t-1})\Bigl[0.06+0.14\{1+\tanh(3Y_{t-1})\}\Bigr]\\
 &+r_{0.40}(Y_{t-2})\Bigl[0.30
 +0.14\frac{\sigma_{t-2}^2}{0.50+\sigma_{t-2}^2}\Bigr]
 +0.08\sigma_{t-1}^2+0.30\sigma_{t-2}^2.
\end{align*}
}
For an estimation date, let $u\in[0,1]$ denote its relative position in the
sample, and define $v=(u-0.30)/0.60$ over the transition interval. The
transition weight $w(u)$ equals $0$ for $u\leq0.30$, $v^2(3-2v)$ for
$0.30<u<0.90$, and $1$ for $u\geq0.90$.
Suppressing the common argument $\Xtrue{t}$, the three
time-varying maps are
{\small
\[
\begin{aligned}
 f_u^{\mathrm{scale}}&=\{0.65+0.80w(u)\}f_-,\qquad
 f_u^{\mathrm{change}}=\{1-w(u)\}f_-+w(u)f_+,\\
 f_u^{\mathrm{joint}}&=\{0.65+0.80w(u)\}
 \bigl[\{1-w(u)\}f_-+w(u)f_+\bigr].
\end{aligned}
\]
}
The transition ends at $u=0.90$, after which the terminal map
applies throughout the forecast segment. All three processes remain bounded.

\subsection{Implementation and evaluation measures}
\label{app:simulation-implementation}

This subsection describes the six estimators in the stationary
experiment and the nine estimators in the time-varying experiment.
\textbf{Kernel} and \textbf{NN-MLP} retain the iterative regression and variance
updates in Section~\ref{sec:algo}.

\textbf{Gaussian Quasi-MLE Benchmarks.} We implement MLE, GJR, and EGARCH
with the \texttt{arch} package. MLE uses a Gaussian GARCH$(p,q)$ model. GJR
uses a Gaussian GJR-GARCH$(p,p,q)$ model, and EGARCH uses a Gaussian
EGARCH$(p,p,q)$ model. Each model includes a constant conditional mean and
uses the true $(p,q)$. The estimated models produce recursive one-step-ahead
variance forecasts over the updating and evaluation periods. If the initial
MLE or GJR optimization does not converge, we repeat estimation from
prespecified interior starting values and retain the converged result with the
largest Gaussian quasi-log-likelihood.

\textbf{Kernel Estimator.} Kernel follows the iterated estimator of
\citet{buhlmann2002algorithm} and uses a positive Gaussian kernel. We initialize the
conditional variances with a Gaussian GARCH$(1,1)$ quasi-MLE. At each of the
$M=8$ iterations, we use inverse estimated-variance weights
$(\widehat\sigma_{t,m-1}^2)^{-1}$ in the kernel regression. We average the last
$K=5$ estimated variance paths and use the averaged path as the variance input
in one final kernel regression. The return and estimated variance inputs enter
the kernel estimator on their original scales without truncation. We select the nearest-neighbor count from
$\{64,128,256,512\}$ and the bandwidth multiplier from
$\{0.35,0.55,0.80,1.10\}$ using the first 80 percent of the estimation sample
for fitting and the final 20 percent for validation. We use the selected pair
throughout all iterations and in the final regression. We prescribe $M=8$ and
$K=5$.

\textbf{LSTM.} This direct benchmark uses the preceding 20
standardized returns and their squares. It has one LSTM layer with 32 hidden
units and a softplus output. Its 4,641 trainable parameters are comparable to
the 4,545--4,929 parameters of NN-MLP across the lag orders
considered. We
minimize squared error for the normalized next squared return using Adam with a
step size of $3\times10^{-3}$, a batch size of 512, and a maximum gradient norm
of 5.
The final 20 percent of the estimation sample selects the epoch, with a
100-epoch cap and patience 10. We then re-estimate the network on the full
estimation sample for the selected number of epochs. Its reported training MSE
uses equal weights.

\textbf{NN-MLP and weighted variants.} The four iterated
NN-MLP estimators use the same input normalization and initial variance path.
Let $\overline{Y^2}$ denote the mean squared return in the estimation sample.
We set
$s_Y=\max\{\sqrt{\overline{Y^2}},0.10\}$,
$B_{\min,T}=\max\{10^{-6},0.01\overline{Y^2}\}$, and
$S_T=\max\{1,4\overline{Y^2}\}$.
We compute all three scaling quantities from the estimation sample. We scale
each return input by $s_Y^{-1}$ and transform each estimated
variance input to
$[(\widehat\sigma^2-B_{\min,T})/(S_T-B_{\min,T})]_{[0,1]}$. The variance
transformation agrees with the bounded variance-input domain used in the
theory. Together, the two transformations place the inputs on comparable
scales and improve optimization stability without changing the estimator's
information set.
We initialize each neural variance path with the mean of the first 100 squared
returns, bounded below by $B_{\min,T}$, and apply the same input
transformations during neural estimation and recursive forecasting.
Each NN-MLP has two hidden ReLU layers with 64 units each. We
use 15 Adam epochs per iteration. The output is the shifted and scaled ReLU
\[
 \widehat f_m(x)=B_{\min,T}+S_T\operatorname{ReLU}\{g_{\theta_m}(x)\}.
\]
The shift enforces positivity, and $S_T$ restores the variance scale. We impose
neither an upper output bound nor the variance-input Lipschitz restriction
$B'<1$ in the numerical experiments.

For Adam, we use a step size of $10^{-3}$ and a maximum gradient norm of 5.
We initialize the network at iteration $m+1$ with the parameter estimates from
iteration $m$, but restart the Adam optimizer.

In the changing-dynamics experiment, NN-Linear uses weights
proportional to $t-r_0$, NN-EW uses half-life $h_T=0.20(T-r_0)$, and NN-Window
uses the most recent $0.30(T-r_0)$ observations. We chose the NN-EW half-life
and NN-Window fraction separately by minimizing the geometric mean of median
future MSEs across the three designs over 20 calibration replications.

\textbf{Stopping rule for NN-MLP iterations.} For weights
$a_t$ on the eligible estimation dates, define the relative path change
\[
 d_{m,a}=\left\{
 \frac{\sum_ta_t(\widehat\sigma_{m,t}^2-
 \widehat\sigma_{m-1,t}^2)^2}
 {\sum_ta_t\widehat\sigma_{m-1,t}^4}
 \right\}^{1/2}.
\]
NN-MLP uses equal weights in this criterion. The extensions use
their own regression weights. Starting at $m=4$, each NN-MLP estimator stops
when $d_{m,a}\leq0.02$ for the first time, with $M\leq8$.

\textbf{Evaluation Measures.}
For weights $a_t$ on the estimation dates with the required lags, the weighted
in-sample variance-path MSE is $\widehat\Delta_{T,M}^{\rm in}(a)
=\sum_t a_t(\widehat\sigma_{t,M,a}^2-\sigma_t^2)^2$. We use the out-of-sample
variance-path MSE defined in Section~\ref{sec_sim} for forecast evaluation.

Tables~\ref{tab:sim-known-insample} and \ref{tab:sim-sample-size-oos}
report MSEs for $T=2000,4000,8000$. The in-sample loss
corresponds to Theorem~\ref{thm:main-upfront}. Within each replication, nested
estimation windows share an endpoint and the same future segments
$(N,H)=(100,1900)$. Estimation and tuning rules remain unchanged.

\begin{table}[!p]
\centering
\renewcommand{\baselinestretch}{1}
\setlength{\tabcolsep}{2pt}
\setlength{\abovecaptionskip}{4pt}
\setlength{\belowcaptionskip}{3pt}
\renewcommand{\arraystretch}{0.94}
\caption{In-sample variance-path MSE by sample size. Medians (Monte Carlo standard deviations) over 50 replications, scaled by 100.}
\label{tab:sim-known-insample}
\fontsize{9}{9}\selectfont
\begin{tabular*}{\textwidth}{@{\extracolsep{\fill}}llrrrrrr@{}}
\toprule
DGP & $T$ & MLE & EGARCH & GJR & Kernel & NN-MLP & LSTM \\
\midrule
\multirow{3}{*}{A$(2,2)$} & 2000 & \textbf{0.38\,{\scriptsize(0.57)}} & 1.30\,{\scriptsize(2.14)} & 0.62\,{\scriptsize(0.63)} & 8.05\,{\scriptsize(9.03)} & 4.45\,{\scriptsize(12.85)} & 2.22\,{\scriptsize(7.89)} \\
 & 4000 & \textbf{0.20\,{\scriptsize(0.20)}} & 1.22\,{\scriptsize(1.26)} & 0.31\,{\scriptsize(0.24)} & 7.11\,{\scriptsize(4.39)} & 2.50\,{\scriptsize(5.47)} & 1.81\,{\scriptsize(3.53)} \\
 & 8000 & \textbf{0.09\,{\scriptsize(0.10)}} & 1.11\,{\scriptsize(0.76)} & 0.16\,{\scriptsize(0.12)} & 5.97\,{\scriptsize(2.28)} & 1.95\,{\scriptsize(2.12)} & 1.54\,{\scriptsize(1.95)} \\
\addlinespace[1.2pt]
\multirow{3}{*}{B$(2,2)$} & 2000 & 1.07\,{\scriptsize(0.27)} & 0.63\,{\scriptsize(0.35)} & \textbf{0.30\,{\scriptsize(0.20)}} & 4.00\,{\scriptsize(1.78)} & 2.11\,{\scriptsize(2.65)} & 0.78\,{\scriptsize(1.94)} \\
 & 4000 & 0.99\,{\scriptsize(0.17)} & 0.55\,{\scriptsize(0.23)} & \textbf{0.17\,{\scriptsize(0.13)}} & 3.43\,{\scriptsize(1.04)} & 1.37\,{\scriptsize(0.99)} & 0.76\,{\scriptsize(1.43)} \\
 & 8000 & 0.97\,{\scriptsize(0.15)} & 0.52\,{\scriptsize(0.22)} & \textbf{0.10\,{\scriptsize(0.06)}} & 2.83\,{\scriptsize(0.87)} & 1.00\,{\scriptsize(1.52)} & 0.74\,{\scriptsize(0.67)} \\
\addlinespace[1.2pt]
\multirow{3}{*}{C$(2,2)$} & 2000 & 2.03\,{\scriptsize(0.57)} & \textbf{0.34\,{\scriptsize(0.26)}} & 0.82\,{\scriptsize(0.74)} & 2.24\,{\scriptsize(0.58)} & 1.36\,{\scriptsize(0.78)} & 0.70\,{\scriptsize(0.97)} \\
 & 4000 & 1.88\,{\scriptsize(0.37)} & \textbf{0.18\,{\scriptsize(0.14)}} & 0.59\,{\scriptsize(0.61)} & 1.69\,{\scriptsize(0.75)} & 0.88\,{\scriptsize(0.55)} & 0.51\,{\scriptsize(0.72)} \\
 & 8000 & 1.78\,{\scriptsize(0.23)} & \textbf{0.07\,{\scriptsize(0.08)}} & 0.55\,{\scriptsize(0.37)} & 1.23\,{\scriptsize(0.33)} & 0.59\,{\scriptsize(0.16)} & 0.36\,{\scriptsize(0.35)} \\
\addlinespace[1.2pt]
\multirow{3}{*}{D$(3,3)$} & 2000 & 8.51\,{\scriptsize(1.22)} & \textbf{3.96\,{\scriptsize(0.88)}} & 7.34\,{\scriptsize(1.79)} & 5.95\,{\scriptsize(2.09)} & 6.62\,{\scriptsize(1.89)} & 5.30\,{\scriptsize(1.42)} \\
 & 4000 & 8.29\,{\scriptsize(0.72)} & \textbf{3.62\,{\scriptsize(0.78)}} & 7.28\,{\scriptsize(1.36)} & 4.76\,{\scriptsize(0.68)} & 4.02\,{\scriptsize(0.63)} & 4.68\,{\scriptsize(5.24)} \\
 & 8000 & 8.27\,{\scriptsize(0.49)} & 3.33\,{\scriptsize(0.47)} & 6.97\,{\scriptsize(0.91)} & 3.76\,{\scriptsize(0.80)} & \textbf{2.81\,{\scriptsize(0.54)}} & 3.65\,{\scriptsize(1.42)} \\
\addlinespace[1.2pt]
\multirow{3}{*}{E$(3,4)$} & 2000 & 9.42\,{\scriptsize(3.04)} & 5.90\,{\scriptsize(2.09)} & 10.07\,{\scriptsize(3.77)} & 5.77\,{\scriptsize(1.38)} & \textbf{5.26\,{\scriptsize(1.79)}} & 5.33\,{\scriptsize(1.54)} \\
 & 4000 & 9.45\,{\scriptsize(2.12)} & 5.44\,{\scriptsize(1.58)} & 9.78\,{\scriptsize(2.70)} & 4.51\,{\scriptsize(1.18)} & \textbf{3.06\,{\scriptsize(0.54)}} & 4.18\,{\scriptsize(1.15)} \\
 & 8000 & 9.42\,{\scriptsize(1.56)} & 5.16\,{\scriptsize(1.14)} & 9.48\,{\scriptsize(2.02)} & 3.46\,{\scriptsize(0.71)} & \textbf{2.12\,{\scriptsize(0.35)}} & 3.28\,{\scriptsize(1.12)} \\
\addlinespace[1.2pt]
\multirow{3}{*}{F$(4,3)$} & 2000 & 12.71\,{\scriptsize(1.15)} & 11.77\,{\scriptsize(1.08)} & 13.19\,{\scriptsize(1.48)} & \textbf{10.11\,{\scriptsize(2.10)}} & 10.29\,{\scriptsize(2.26)} & 12.42\,{\scriptsize(3.84)} \\
 & 4000 & 12.28\,{\scriptsize(0.72)} & 11.10\,{\scriptsize(0.65)} & 12.59\,{\scriptsize(0.75)} & 8.49\,{\scriptsize(1.53)} & \textbf{6.69\,{\scriptsize(1.12)}} & 11.72\,{\scriptsize(1.01)} \\
 & 8000 & 12.08\,{\scriptsize(0.55)} & 10.72\,{\scriptsize(0.42)} & 12.31\,{\scriptsize(0.62)} & 7.16\,{\scriptsize(0.87)} & \textbf{4.99\,{\scriptsize(0.68)}} & 11.16\,{\scriptsize(1.04)} \\
\addlinespace[1.2pt]
\multirow{3}{*}{G$(4,4)$} & 2000 & 10.28\,{\scriptsize(1.87)} & \textbf{5.10\,{\scriptsize(2.33)}} & 13.97\,{\scriptsize(7.87)} & 7.17\,{\scriptsize(1.46)} & 6.03\,{\scriptsize(2.58)} & 6.37\,{\scriptsize(2.96)} \\
 & 4000 & 10.35\,{\scriptsize(1.31)} & 4.38\,{\scriptsize(1.49)} & 14.45\,{\scriptsize(4.68)} & 5.90\,{\scriptsize(1.06)} & \textbf{3.58\,{\scriptsize(0.76)}} & 5.89\,{\scriptsize(0.88)} \\
 & 8000 & 10.17\,{\scriptsize(0.88)} & 4.34\,{\scriptsize(1.10)} & 13.39\,{\scriptsize(3.32)} & 4.66\,{\scriptsize(0.55)} & \textbf{2.30\,{\scriptsize(0.45)}} & 5.61\,{\scriptsize(2.59)} \\
\addlinespace[1.2pt]
\multirow{3}{*}{H$(5,5)$} & 2000 & 4.22\,{\scriptsize(0.45)} & 4.43\,{\scriptsize(0.46)} & 4.63\,{\scriptsize(0.59)} & \textbf{4.01\,{\scriptsize(1.47)}} & 6.42\,{\scriptsize(1.48)} & 4.18\,{\scriptsize(0.50)} \\
 & 4000 & 3.99\,{\scriptsize(0.20)} & 3.71\,{\scriptsize(0.21)} & 4.14\,{\scriptsize(0.25)} & \textbf{3.22\,{\scriptsize(1.07)}} & 3.53\,{\scriptsize(0.76)} & 3.93\,{\scriptsize(0.36)} \\
 & 8000 & 3.82\,{\scriptsize(0.14)} & 3.51\,{\scriptsize(0.13)} & 3.93\,{\scriptsize(0.16)} & 2.56\,{\scriptsize(0.56)} & \textbf{2.16\,{\scriptsize(0.28)}} & 3.63\,{\scriptsize(0.73)} \\
\bottomrule
\end{tabular*}
\par\vspace{9pt}
\caption{Out-of-sample variance-path MSE by sample size. Medians (Monte Carlo standard deviations) over 50 replications, scaled by 100.}
\label{tab:sim-sample-size-oos}
\fontsize{9}{9}\selectfont
\begin{tabular*}{\textwidth}{@{\extracolsep{\fill}}llrrrrrr@{}}
\toprule
DGP & $T$ & MLE & EGARCH & GJR & Kernel & NN-MLP & LSTM \\
\midrule
\multirow{3}{*}{A$(2,2)$} & 2000 & \textbf{0.30\,{\scriptsize(0.36)}} & 1.17\,{\scriptsize(1.49)} & 0.60\,{\scriptsize(0.37)} & 7.71\,{\scriptsize(5.86)} & 4.61\,{\scriptsize(6.25)} & 2.25\,{\scriptsize(4.35)} \\
 & 4000 & \textbf{0.18\,{\scriptsize(0.22)}} & 0.86\,{\scriptsize(1.80)} & 0.30\,{\scriptsize(0.34)} & 5.73\,{\scriptsize(5.87)} & 2.56\,{\scriptsize(4.13)} & 1.40\,{\scriptsize(3.49)} \\
 & 8000 & \textbf{0.08\,{\scriptsize(0.12)}} & 0.72\,{\scriptsize(1.55)} & 0.15\,{\scriptsize(0.16)} & 4.35\,{\scriptsize(5.12)} & 1.62\,{\scriptsize(2.15)} & 1.07\,{\scriptsize(2.99)} \\
\addlinespace[1.2pt]
\multirow{3}{*}{B$(2,2)$} & 2000 & 1.09\,{\scriptsize(0.35)} & 0.79\,{\scriptsize(0.64)} & \textbf{0.29\,{\scriptsize(0.24)}} & 4.35\,{\scriptsize(2.15)} & 2.40\,{\scriptsize(1.29)} & 1.07\,{\scriptsize(1.92)} \\
 & 4000 & 0.98\,{\scriptsize(0.25)} & 0.56\,{\scriptsize(0.49)} & \textbf{0.18\,{\scriptsize(0.14)}} & 3.43\,{\scriptsize(1.95)} & 1.46\,{\scriptsize(1.00)} & 0.86\,{\scriptsize(2.38)} \\
 & 8000 & 0.93\,{\scriptsize(0.21)} & 0.49\,{\scriptsize(0.32)} & \textbf{0.09\,{\scriptsize(0.06)}} & 2.65\,{\scriptsize(1.53)} & 0.86\,{\scriptsize(1.27)} & 0.71\,{\scriptsize(0.72)} \\
\addlinespace[1.2pt]
\multirow{3}{*}{C$(2,2)$} & 2000 & 1.97\,{\scriptsize(0.45)} & \textbf{0.31\,{\scriptsize(0.31)}} & 0.88\,{\scriptsize(0.70)} & 2.08\,{\scriptsize(0.82)} & 1.37\,{\scriptsize(0.67)} & 0.71\,{\scriptsize(1.07)} \\
 & 4000 & 1.89\,{\scriptsize(0.31)} & \textbf{0.17\,{\scriptsize(0.12)}} & 0.60\,{\scriptsize(0.47)} & 1.72\,{\scriptsize(0.70)} & 0.89\,{\scriptsize(0.39)} & 0.54\,{\scriptsize(0.66)} \\
 & 8000 & 1.77\,{\scriptsize(0.22)} & \textbf{0.08\,{\scriptsize(0.08)}} & 0.51\,{\scriptsize(0.46)} & 1.23\,{\scriptsize(0.44)} & 0.57\,{\scriptsize(0.22)} & 0.41\,{\scriptsize(0.28)} \\
\addlinespace[1.2pt]
\multirow{3}{*}{D$(3,3)$} & 2000 & 8.55\,{\scriptsize(1.00)} & \textbf{4.15\,{\scriptsize(1.05)}} & 7.56\,{\scriptsize(1.92)} & 5.99\,{\scriptsize(1.95)} & 7.07\,{\scriptsize(1.81)} & 5.58\,{\scriptsize(1.41)} \\
 & 4000 & 8.22\,{\scriptsize(0.79)} & \textbf{3.57\,{\scriptsize(0.77)}} & 7.27\,{\scriptsize(1.36)} & 4.80\,{\scriptsize(0.93)} & 4.17\,{\scriptsize(1.37)} & 4.73\,{\scriptsize(6.92)} \\
 & 8000 & 8.04\,{\scriptsize(0.71)} & 3.28\,{\scriptsize(0.53)} & 6.89\,{\scriptsize(1.25)} & 3.80\,{\scriptsize(0.96)} & \textbf{2.92\,{\scriptsize(0.57)}} & 3.66\,{\scriptsize(1.38)} \\
\addlinespace[1.2pt]
\multirow{3}{*}{E$(3,4)$} & 2000 & 9.65\,{\scriptsize(2.80)} & 6.29\,{\scriptsize(3.34)} & 10.23\,{\scriptsize(3.85)} & 5.93\,{\scriptsize(1.62)} & 5.63\,{\scriptsize(1.46)} & \textbf{5.24\,{\scriptsize(1.72)}} \\
 & 4000 & 9.30\,{\scriptsize(2.38)} & 5.21\,{\scriptsize(2.22)} & 9.43\,{\scriptsize(2.87)} & 4.71\,{\scriptsize(1.34)} & \textbf{3.21\,{\scriptsize(0.72)}} & 4.24\,{\scriptsize(1.34)} \\
 & 8000 & 8.93\,{\scriptsize(2.01)} & 4.86\,{\scriptsize(1.58)} & 9.00\,{\scriptsize(2.26)} & 3.59\,{\scriptsize(0.97)} & \textbf{2.14\,{\scriptsize(0.40)}} & 3.26\,{\scriptsize(1.02)} \\
\addlinespace[1.2pt]
\multirow{3}{*}{F$(4,3)$} & 2000 & 12.90\,{\scriptsize(1.20)} & 12.04\,{\scriptsize(1.34)} & 13.86\,{\scriptsize(1.43)} & 10.96\,{\scriptsize(2.41)} & \textbf{10.86\,{\scriptsize(2.50)}} & 12.49\,{\scriptsize(5.19)} \\
 & 4000 & 12.32\,{\scriptsize(1.02)} & 11.11\,{\scriptsize(0.90)} & 12.89\,{\scriptsize(1.08)} & 9.03\,{\scriptsize(1.93)} & \textbf{6.85\,{\scriptsize(1.27)}} & 11.90\,{\scriptsize(1.21)} \\
 & 8000 & 12.09\,{\scriptsize(0.99)} & 10.84\,{\scriptsize(0.81)} & 12.53\,{\scriptsize(1.00)} & 7.59\,{\scriptsize(1.22)} & \textbf{5.13\,{\scriptsize(0.76)}} & 11.64\,{\scriptsize(1.63)} \\
\addlinespace[1.2pt]
\multirow{3}{*}{G$(4,4)$} & 2000 & 10.94\,{\scriptsize(1.44)} & \textbf{5.30\,{\scriptsize(2.50)}} & 14.14\,{\scriptsize(8.42)} & 7.96\,{\scriptsize(1.78)} & 6.60\,{\scriptsize(2.78)} & 6.77\,{\scriptsize(2.98)} \\
 & 4000 & 10.59\,{\scriptsize(1.20)} & 4.82\,{\scriptsize(1.89)} & 14.48\,{\scriptsize(6.38)} & 6.09\,{\scriptsize(1.61)} & \textbf{3.65\,{\scriptsize(0.92)}} & 6.03\,{\scriptsize(1.12)} \\
 & 8000 & 10.36\,{\scriptsize(1.09)} & 4.65\,{\scriptsize(1.42)} & 14.51\,{\scriptsize(5.33)} & 4.92\,{\scriptsize(0.91)} & \textbf{2.31\,{\scriptsize(0.54)}} & 5.69\,{\scriptsize(4.10)} \\
\addlinespace[1.2pt]
\multirow{3}{*}{H$(5,5)$} & 2000 & 4.24\,{\scriptsize(0.42)} & 4.31\,{\scriptsize(0.47)} & 4.52\,{\scriptsize(0.49)} & \textbf{3.97\,{\scriptsize(1.46)}} & 6.45\,{\scriptsize(1.46)} & 4.17\,{\scriptsize(0.46)} \\
 & 4000 & 3.86\,{\scriptsize(0.29)} & 3.73\,{\scriptsize(0.27)} & 4.06\,{\scriptsize(0.28)} & \textbf{3.27\,{\scriptsize(1.02)}} & 3.59\,{\scriptsize(0.86)} & 3.88\,{\scriptsize(0.41)} \\
 & 8000 & 3.77\,{\scriptsize(0.25)} & 3.47\,{\scriptsize(0.21)} & 3.88\,{\scriptsize(0.26)} & 2.61\,{\scriptsize(0.47)} & \textbf{2.21\,{\scriptsize(0.29)}} & 3.60\,{\scriptsize(0.66)} \\
\bottomrule
\end{tabular*}
\end{table}

\FloatBarrier

\section{Real-data details}
\label{app:real-reference}

\noindent\textbf{Daily data.}
\revadd{We obtain adjusted daily prices and five-minute records from Alpaca
Markets' SIP feed. To compute BDS statistics and select stocks, we fit
zero-mean Gaussian GARCH, GJR, and EGARCH models with $p,q\in\{1,2,3,4\}$
to returns from 2016 through December 28, 2021, for eligible S\&P 100 stocks.
For each stock, we divide returns by the fitted volatility from the
converged model with the lowest Bayesian information criterion (BIC). We
compute the BDS statistic with embedding dimension two and a threshold of
half the standardized-return standard deviation.}
\revadd{Table~\ref{tab:real-selection} reports the eight stocks with the highest
BDS statistics among those with at least 90\% daily realized-variance coverage
in each study period. The table also gives each stock's most correlated panel
member based on 2016--2021 daily returns.}
\revadd{Let $r_{t,\mathrm{overnight}}$ denote the return from the previous close
to day $t$'s opening price, and $r_{t,j,\mathrm{5min}}$ its five-minute returns.
Daily realized variance is
$RV_t=r_{t,\mathrm{overnight}}^2+\sum_j r_{t,j,\mathrm{5min}}^2$.
We fit the models to squared daily returns and score them against $RV_t$.}
\revadd{We choose lag orders for the parametric and Kernel methods by QLIKE
over the December 29, 2021--December 28, 2023 validation period, using models
fitted before that period.
NN-MLP and NN-EW use a constant initial variance path, two 64-unit layers,
and validation-selected settings. After selection, we refit all methods
on the 2,011 daily returns
from January 2016 through December 2023 and evaluate them on 502 observations
from January 2024 through December 2025.}

\begin{table}[ht!]
{
\centering
\caption{Stock selection for daily data. Auxiliary stocks are selected by
daily-return correlations over 2016--2021.}
\label{tab:real-selection}
\small
\setlength{\tabcolsep}{5pt}
\begin{tabular}{@{}l*{8}{c}@{}}
\toprule
 & MMM & UNP & IBM & NKE & NEE & TGT & SBUX & KHC\\
\midrule
BDS & 4.1144 & 3.4150 & 3.2088 & 2.9916 & 2.8526 & 2.8237 & 2.7894 & 2.7473\\
Auxiliary & IBM & MMM & MMM & SBUX & SBUX & NKE & NKE & MMM\\
Correlation & 0.5694 & 0.5516 & 0.5694 & 0.5135 & 0.4128 & 0.3494 & 0.5135 & 0.3939\\
\bottomrule
\end{tabular}
}
\end{table}
\FloatBarrier

\par\noindent\textbf{Option-price data.}
\revadd{We obtain option prices from Alpaca Markets' Basic indicative feed\footnote{Alpaca Markets,
\href{https://docs.alpaca.markets/us/docs/historical-option-data}{``Historical Option Data,''}
accessed August 29, 2026.}, and VIX and the three-month Treasury rate from
FRED series VIXCLS and DGS3MO. Training covers February 5, 2024--June 30,
2025 (351 days), validation July--December 2025 (128), and testing
January--August 21, 2026 (160).}
\revadd{To construct implied volatility, we invert Black's pricing formula
using the call and put prices, strike, time to expiry, parity-implied forward
price, and three-month Treasury rate. We match calls and puts by strike and
expiry, select the 20--45-day expiry closest to 30 days and the strike
closest to the median parity-implied forward, and average the valid call and
put solutions. For M2 with the $Y_t^2$ response, we use the auxiliary stock's
lagged five-day log implied-volatility change. For the $RV_t$ response, we use
the residual from a training-sample regression of its lagged daily log change
on target implied-volatility and VIX changes, with an intercept.}
\revadd{For each stock, method, and response, we select a configuration by
mean validation QLIKE across M0--M2 and fix it across information sets.
After refitting through 2025, we initialize lags on five test days and score
155 days.}

\FloatBarrier

\par\noindent\textbf{Five-minute data.}
\revadd{For MMM and UNP, we obtain intraday data from the Alpaca SIP feed. Training,
validation, and test cover
Jan.~3, 2022--May~21, 2024 (595 days), May~22, 2024--Mar.~12, 2025 (198),
and Mar.~13--Dec.~31, 2025 (200), respectively.}
\revadd{For each five-minute interval, we compute returns from midpoint prices
and use the sum of five squared one-minute midpoint returns as the regression
response. To remove intraday scale patterns, we compute mean squared returns
by five-minute slot on training days and normalize the factors to average one.
For fitting, we divide returns by the square root of the corresponding factor
and realized variance by the factor. We multiply forecasts by that factor
before scoring.}
\revadd{The augmented information set adds
realized variances from the previous two intervals, the auxiliary stock's
preceding return, trading activity, and time of day. The two trading-activity
regressors are the number of midpoint-price changes and the average seconds
between changes in the previous five-minute interval. We apply $\log(1+x)$
to each regressor.}
\revadd{We select configurations by validation QLIKE for each stock, method,
and information set. Neural forecasts average three independently initialized
networks for validation and five for testing. The first five intervals of each
test day initialize lag histories, leaving 72 scored forecasts.}
\FloatBarrier

\setlength{\bibsep}{7pt plus 3pt minus 2pt}
\begingroup
\let\originalbibitem\bibitem
\renewcommand{\bibitem}[2][]{%
  \par\spaceskip=0pt\relax
  \originalbibitem[#1]{#2}%
  \ifstrequal{#2}{lu2020deep}{\spaceskip=3.3pt plus 1.5pt minus 1pt\relax}{}%
  \ifstrequal{#2}{meister2016statistical}{\spaceskip=3pt plus 1.5pt minus 1pt\relax}{}%
  \ignorespaces
}

\endgroup

\clearpage

\setcounter{section}{0}
\setcounter{subsection}{0}
\setcounter{equation}{0}
\setcounter{theorem}{0}
\setcounter{lemma}{0}
\setcounter{proposition}{0}
\setcounter{corollary}{0}
\setcounter{assumption}{0}
\setcounter{definition}{0}
\setcounter{model}{0}
\setcounter{remark}{0}
\setcounter{table}{0}
\setcounter{figure}{0}

\renewcommand{\thesection}{S.\arabic{section}}
\renewcommand{\theequation}{S.\arabic{equation}}
\renewcommand{\thetheorem}{S.\arabic{theorem}}
\renewcommand{\thelemma}{S.\arabic{lemma}}
\renewcommand{\theproposition}{S.\arabic{proposition}}
\renewcommand{\thecorollary}{S.\arabic{corollary}}
\renewcommand{\theassumption}{S.\arabic{assumption}}
\renewcommand{\thedefinition}{S.\arabic{definition}}
\renewcommand{\themodel}{S.\arabic{model}}
\renewcommand{\theremark}{S.\arabic{remark}}
\renewcommand{\thetable}{S.\arabic{table}}
\renewcommand{\thefigure}{S.\arabic{figure}}

\vspace*{18pt}
\begin{center}
{\Large Supplementary Materials for\\[2pt]
\emph{Nonparametric GARCH: An Iterated Neural Estimator}\par}
\vspace{14pt}
Ruizhi Deng\textsuperscript{a}, Guohao Shen\textsuperscript{a},
Ngai Hang Chan\textsuperscript{b}\par
\vspace{8pt}
{\small\itshape
\textsuperscript{a}Department of Applied Mathematics,
The Hong Kong Polytechnic University\\
\textsuperscript{b}Department of Biostatistics,
City University of Hong Kong\par}
\end{center}
\vspace{16pt}

This supplement provides proofs of the results in the main text.
Section~\ref{app:net} describes the network configurations and their
approximation and complexity properties.
Sections~\ref{app:insample} and~\ref{app:lower} prove in-sample upper and
minimax lower bounds.  Sections~\ref{app:population-transfer} and~\ref{app:apt-rates}
prove bounds for recursion-map estimation error and out-of-sample variance-forecast
error. Section~\ref{app:simulation-results}
reports an additional simulation comparison.

Throughout the upper-bound proofs, we keep the fitted orders $p\geq\pstar$ and
$q\geq\qstar$ constant as $T$ varies, and set $r_0=\max(p,q)$ and $n_T=T-r_0$.
For comparison with fitted networks, $\ftrue$ also denotes its extension
to the fitted input space that ignores redundant lag coordinates.

\section{Network approximation and proof preliminaries}
\label{app:def-ass}\label{app:net}

We bound the network approximation error for $\ftrue$ and the covering number
of the network class. Because return inputs are unbounded, we establish
these bounds on a growing input domain and show that the observed return
coordinates fall within it with high probability.

\begin{definition}[H\"older ball]
\label{def:holder}
Write $\hol=k_{\hol}+\alpha_{\hol}$, where
$k_{\hol}=\lceil\hol\rceil-1$ and
$\alpha_{\hol}=\hol-k_{\hol}\in(0,1]$.  With the usual multi-index notation,
define
\[
 \|f\|_{\mathcal H^\hol}
 =\max_{|a|\leq k_{\hol}}\|\partial^a f\|_\infty
  +\max_{|a|=k_{\hol}}\sup_{x\neq x'}
   \frac{|\partial^a f(x)-\partial^a f(x')|}
        {\|x-x'\|_2^{\alpha_{\hol}}}.
\]
Then $\mathcal H^\hol(\R^d,B)=\{f:\R^d\to\R:
\|f\|_{\mathcal H^\hol}\leq B\}$.
\end{definition}

Definition~\ref{def:holder} makes the smoothness condition in
Assumption~\ref{ass:hol-up} precise.  Together,
Assumptions~\ref{ass:innov-up} and~\ref{ass:hol-up} imply
$\ftrue(x)\in[\Bmin,\Bmax]$ on $\R^{\fdim}$.  We therefore project every
network output onto this interval.  The projection keeps the generated
variance inputs in the same interval and does not increase either the
approximation error or the variance-input Lipschitz constant.

Because the return inputs need not be bounded, we analyze them on a growing
domain.  For fitted orders $(p,q)$, restrict the return coordinates to
$[-C_Y\sqrt{\log T},C_Y\sqrt{\log T}]^p$, where $C_Y$ is sufficiently large.
The variance coordinates remain in $[\Bmin,\Bmax]^q$.  Denote the product of
these two sets by $\mathcal X_T$.  The sub-Gaussian innovation condition and
the global range bound imply that, for any $A>0$ and sufficiently large $C_Y$,
\begin{equation}
 \Prob\left(\max_{1\leq t\leq T}|Y_t|>
 C_Y\sqrt{\log T}\right)\leq T^{-A}.
 \label{eq:apt-return-truncation}
\end{equation}
This follows from $|Y_t|\leq\sqrt{\Bmax}|\varepsilon_t|$ and a union bound.
Thus the return-coordinate restriction holds at all observed dates with high
probability.

Let $\Fclass_T$ be a compactly parameterized class of projected ReLU networks
of the form specified by \eqref{eq:apt-mlp} in the main text that satisfy
$|f(y,s)-f(y,s')|\leq B'\lVert s-s'\rVert_\infty$ for all return inputs $y$
and all $s,s'\in[\Bmin,\Bmax]^q$.
For $\epsilon>0$, the covering number
$\mathcal N(\epsilon,\Fclass_T,\|\cdot\|_{\infty,\mathcal X_T})$ is the
minimum number of networks in $\Fclass_T$ needed to approximate every
member within sup-norm error $\epsilon$ on $\mathcal X_T$.

The next proposition verifies the approximation and covering-number
properties of this class.

\begin{proposition}[Approximation and complexity of the network class]
\label{prop:network-up}
Suppose Assumptions~\ref{ass:innov-up}--\ref{ass:lip-up} hold and $B'>\Dcon$.
For $2\leq\snet\leq T^{C_S}$, where $C_S<\infty$ does not depend on $T$, one
can choose a network structure, a compact parameter set, and a constant
$C_{\rm arch}<\infty$
such that
$1\leq\dnet\leq C_{\rm arch}\log(2+\snet)$,
$\wnet\leq C_{\rm arch}\snet$.  The resulting class $\Fclass_T$ contains a
deterministic comparison network $f_{\snet}$, and there are constants
$C>0$ and $c\geq0$, independent of $T$ and $\snet$, such that
\begin{align}
 \|f_{\snet}-\ftrue\|_{\infty,\mathcal X_T}^2
 &\leq C(\log T)^c\snet^{-2\hol/(\fdim)},
 \label{eq:apt-approx}\\
 \log\mathcal N(\epsilon,\Fclass_T,
 \|\cdot\|_{\infty,\mathcal X_T})
 &\leq C\snet\dnet\log(CT/\epsilon),\qquad 0<\epsilon\leq1.
 \label{eq:apt-entropy}
\end{align}
When $p\geq\pstar$ and $q\geq\qstar$, the approximating network $f_{\snet}$
assigns zero weights to redundant inputs.  Its approximation rate therefore
depends on the intrinsic dimension $\pstar+\qstar$.  Constants may depend on
$B'-\Dcon$.
\end{proposition}

\begin{proof}
The proof has two parts.  We first construct an approximating network that
satisfies both the approximation bound and the variance-input Lipschitz bound.  We then
verify the covering-number bound.
To construct the approximating network, we clip returns to their range in $\mathcal X_T$, ignore
redundant lags, and rescale the structural return coordinates to
$[0,1]^{\pstar}$.  Let $\bar f_T$ denote $\ftrue$ in
these rescaled coordinates.  It has H\"older radius
$B_T\leq C(\log T)^{c_0}$, while Assumption~\ref{ass:lip-up} gives
$|\bar f_T(z,s)-\bar f_T(z,s')|\leq\Dcon\|s-s'\|_\infty$.

If $\hol>1$, we partition the intrinsic input domain into cells of side
$K^{-1}$.  For each cell $i$, let $P_i$ be the degree-$k_{\hol}$ Taylor
polynomial of $\bar f_T$ centered there.  Choose nonnegative partition-of-unity
functions $\phi_i(z,s)$
supported within $C/K$ of cell $i$ and satisfying
$\sum_i\phi_i(z,s)=1$ on the domain \citep[Lemma~C.3]{guhring2020error}.
Set $P_K(z,s)=\sum_i\phi_i(z,s)P_i(z,s)$.
On the support of $\phi_i$, Assumption~\ref{ass:hol-up} bounds the value and
first-derivative errors of $P_i$ by $CB_TK^{-\hol}$ and
$CB_TK^{-(\hol-1)}$.  The weights have bounded overlap and first derivatives
of order $K$. Hence $P_K$ satisfies the same error bounds.

The construction in \citet[proof of Lemma~C.6, using Lemmas~C.3
and C.5]{guhring2020error} provides a ReLU network $g_K$ of depth at
most $C\log(2+K)$ and width at most $CK^{\fdim}\log(2+K)$ that approximates
$P_K$ with
\[
 \|g_K-P_K\|_\infty\leq CB_TK^{-\hol},\qquad
 \|\nabla(g_K-P_K)\|_{L^\infty}\leq CB_TK^{-(\hol-1)}.
\]
We interpret the derivatives of the ReLU network almost everywhere.
Combining the network and Taylor errors gives
\[
 \|g_K-\bar f_T\|_\infty\leq CB_TK^{-\hol},\qquad
 \operatorname{Lip}_s(g_K)\leq\Dcon+CB_TK^{-(\hol-1)}.
\]
This construction uses only the prescribed polynomial degree and coefficient
bounds of order $B_T$.  Theorem~4.1 of \citet{guhring2020error} assumes
integer-order Sobolev smoothness for its polynomial approximation step,
which the H\"older Taylor bounds above replace here.
Thus $\operatorname{Lip}_s(g_K)\leq B'$ when
$K^{\hol-1}\geq CB_T/(B'-\Dcon)$.

If $0<\hol\leq1$, the preceding derivative approximation does not apply
because $\bar f_T$ need not be differentiable in its return inputs.
Instead, we use a finite grid to construct an approximation whose
variance-input Lipschitz constant remains below $B'$.
We set $B_0=(\Dcon+B')/2$. Then $\Dcon<B_0<B'$. Let $(z_i)$ be a return grid of
mesh $K^{-1}$ and $\mathcal G_K$ a variance grid of mesh at most
$K^{-\hol}$.  At each return-grid point $z_i$, define
\[
 r_i(s)=\min_{u\in\mathcal G_K}
 \{\bar f_T(z_i,u)+B_0\|s-u\|_\infty\}.
\]
For $\qstar>0$, each map
$s\mapsto\bar f_T(z_i,u)+B_0\|s-u\|_\infty$ for a given $u\in\mathcal G_K$
is $B_0$-Lipschitz, as is their pointwise minimum $r_i$.
Since $B_0>\Dcon$ and $\bar f_T(z_i,\cdot)$ is $\Dcon$-Lipschitz, $r_i$
agrees with it on the variance grid and lies above it elsewhere.
A grid point within $K^{-\hol}$ of $s$ gives
$0\leq r_i(s)-\bar f_T(z_i,s)\leq(B_0+\Dcon)K^{-\hol}$.
When~$\qstar=0$, set $r_i=\bar f_T(z_i)$.  In both cases, $r_i$ has an exact
ReLU representation because its definition uses only constants, absolute
values, and finite maxima and minima.  For each return input $z$, we take a
weighted average of the $r_i(s)$ from nearby grid points $z_i$.  The weights
are nonnegative, depend only on $z$, and sum to one.  Thus the average remains
$B_0$-Lipschitz in $s$.  We approximate this average using the ReLU
partition and multiplication modules of
\citet[Lemmas~C.3 and C.5]{guhring2020error}, taking multiplication accuracy
of order $K^{-(\hol+1)}$.  The grid error above and H\"older continuity in
the return coordinates then give, for sufficiently large $K$, a network
$g_K$ satisfying
\[
 \|g_K-\bar f_T\|_\infty\leq CB_TK^{-\hol},\qquad
 \operatorname{Lip}_s(g_K)\leq B_0+CK^{-\hol}\leq B'.
\]
The grids use $O(K^{\pstar+\hol\qstar})\leq O(K^{\fdim})$ points.

For either range of $\hol$, the resulting $g_K$ has depth at most
$C\log(2+K)$, width at most $CK^{\fdim}\log(2+K)$, and at most
$CK^{\fdim}\{\log(2+K)\}^2$ parameters, with weights bounded by $CB_TK^C$
\citep{guhring2020error}.  Taking
$\snet\asymp K^{\fdim}\{\log(2+K)\}^2$ and substituting this relation
into the preceding approximation bound gives \eqref{eq:apt-approx} for the
projected network.  If the parameter budget is too small for this
construction to satisfy the Lipschitz bound, the constant network equal to
$\Bmin$ also satisfies \eqref{eq:apt-approx} because $\ftrue$ is bounded and such budgets
grow at most as a power of $\log T$.

To bound the covering number, we choose a parameter box with polynomial
radius in $T$ for this network architecture. The sup-norm covering argument of
\citet[Theorem~3.5]{shen2024exploring} then gives \eqref{eq:apt-entropy} on
$\mathcal X_T$.  Output projection and restriction to the closed
$B'$-Lipschitz subclass preserve this bound up to a constant change in the
covering radius. Finally, the empirical criterion is Borel measurable in the data and
preceding parameters and continuous in the current parameter. It therefore
admits a measurable minimizer at each iteration
\citep[Theorem~18.19]{aliprantis2006infinite}.
\end{proof}

\section{Proof of Theorem~\ref{thm:main-upfront}}
\label{app:insample}

Because every iteration uses the same observations, the variance inputs at
stage $m$ depend on networks estimated from the same sample. We first
separate the in-sample error into approximation, error-propagation, and
stochastic terms. Proposition~\ref{prop:network-up} controls the approximation
error. We bound the stochastic term uniformly over deterministic sequences
of candidate networks before applying the bound to the estimated sequence.
Contraction then gives an error recursion, and balancing approximation and
sampling errors yields the stated rate.

\begin{proof}[Proof of Theorem~\ref{thm:main-upfront}]
\noindent\textbf{Error decomposition and approximation.}\par\nobreak
Write $P_Tv=n_T^{-1}\sum_{t=r_0+1}^T v_t$ and
$\Delta_{T,m}^{\rm in}=\E P_T(\volsq_t-\volhat{t,m})^2$ for
$m=0,\ldots,M$. Let
$X_{t,m-1}=(Y_{t-1:t-p},\volhat{t-1:t-q,m-1})$ denote the generated
regressor at iteration $m\geq1$.
Let $f_{\snet}\in\Fclass_T$ be the deterministic comparison network from
Proposition~\ref{prop:network-up}. At iteration $m$, the least-squares estimator
$\widehat f_m$ minimizes the empirical squared loss over $\Fclass_T$.
Its loss is therefore no larger than that of $f_{\snet}$. Using
$\volhat{t,m}=\widehat f_m(X_{t,m-1})$, we obtain, for $1\leq m\leq M$,
\[
 P_T\{Y_t^2-\volhat{t,m}\}^2
 \leq P_T\{Y_t^2-f_{\snet}(X_{t,m-1})\}^2.
\]
Substituting $Y_t^2=\volsq_t+\mds_t$ and expanding both squares cancels
the common term $P_T\mds_t^2$. Taking expectations then yields
\[
 \Delta_{T,m}^{\rm in}
 \leq\E P_T\{\volsq_t-f_{\snet}(X_{t,m-1})\}^2
 +2\E P_T\left[\mds_t
 \{\widehat f_m(X_{t,m-1})-f_{\snet}(X_{t,m-1})\}\right].
\]

We next separate the first term into network approximation error and error
in the preceding variance estimates. To apply the approximation bound on
$\mathcal X_T$, we use the return-truncation inequality
\eqref{eq:apt-return-truncation}. For any $A>2$, choosing $C_Y$ sufficiently
large ensures that all generated regressors lie in $\mathcal X_T$ with
probability at least $1-T^{-A}$. On this event, we add and subtract
$\ftrue(X_{t,m-1})$ inside the squared difference. Young's inequality and
Assumption~\ref{ass:lip-up} then give, for $\tau>0$,
\begin{equation}
\begin{aligned}
 \{\volsq_t-f_{\snet}(X_{t,m-1})\}^2
 \leq{}&(1+\tau)\Dcon^2\sum_{j=1}^{\qstar}\omega_j
 (\volsq_{t-j}-\volhat{t-j,m-1})^2\\
 &+C_\tau\|f_{\snet}-\ftrue\|_{\infty,\mathcal X_T}^2.
\end{aligned}
 \label{eq:apt-direct-comparator}
\end{equation}
To compare the lagged error average with $\Delta_{T,m-1}^{\rm in}$, substitute
$s=t-j$. For $1\leq j\leq\qstar$,
\begin{align*}
 &\E P_T(\volsq_{t-j}-\volhat{t-j,m-1})^2
 =\Delta_{T,m-1}^{\rm in}
 +\frac{1}{n_T}\sum_{s=r_0+1-j}^{r_0}
 \E(\volsq_s-\volhat{s,m-1})^2\\
 &\qquad-\frac{1}{n_T}\sum_{s=T-j+1}^{T}
 \E(\volsq_s-\volhat{s,m-1})^2
 \leq\Delta_{T,m-1}^{\rm in}
 +\frac{j(\Bmax-\Bmin)^2}{n_T}.
\end{align*}
The subtracted terminal sum is nonnegative. The initial sum contains $j$
terms, each bounded by $(\Bmax-\Bmin)^2$. Since the weights are positive and
sum to one, weighted averaging gives
\[
 \sum_{j=1}^{\qstar}\omega_j\E P_T
 (\volsq_{t-j}-\volhat{t-j,m-1})^2
 \leq\Delta_{T,m-1}^{\rm in}
 +\frac{C}{n_T}\sum_{j=1}^{\qstar}j\omega_j
 \leq\Delta_{T,m-1}^{\rm in}+\frac{C\qstar}{n_T}.
\]
Boundedness of the comparison network $f_{\snet}$ also gives
\begin{align*}
 &\E P_T\left[
 \{\volsq_t-f_{\snet}(X_{t,m-1})\}^2
 \,\mathbf 1_{\{X_{t,m-1}\notin\mathcal X_T\}}\right]\\
 &\qquad\leq C\Prob\left(\max_{1\leq s\leq T}|Y_s|>
 C_Y\sqrt{\log T}\right)\leq CT^{-A}.
\end{align*}
Together with \eqref{eq:apt-direct-comparator}, these inequalities bound the
first term in the loss expansion. Substitution gives
\begin{align*}
 \Delta_{T,m}^{\rm in}
 \leq{}&(1+\tau)\Dcon^2\Delta_{T,m-1}^{\rm in}
 +C_\tau\|f_{\snet}-\ftrue\|_{\infty,\mathcal X_T}^2\\
 &+2\E P_T\left[\mds_t
 \{\widehat f_m(X_{t,m-1})-f_{\snet}(X_{t,m-1})\}\right]
 +C_\tau\left(\frac{\qstar}{n_T}+T^{-A}\right).
\end{align*}
The first term propagates the preceding variance error. The second is the
network approximation error, and the third is the stochastic term. The
remainder accounts for lag boundaries and return tails. When $\qstar=0$,
the propagation and lag-boundary terms disappear.
Proposition~\ref{prop:network-up} gives
$\|f_{\snet}-\ftrue\|_{\infty,\mathcal X_T}^2
\leq C(\log T)^c\snet^{-2\hol/(\fdim)}$.
It remains to control the stochastic term.

\smallskip
\noindent\textbf{Uniform control of the stochastic term.}\par\nobreak
The estimated networks depend on the full sample. Their evaluations at a given
date may therefore depend on later returns. To bound the stochastic term, we first consider
deterministic $c\in[\Bmin,\Bmax]$ and $g_r\in\Fclass_T$,
$r=1,\ldots,M$. We write $g_{1:m}=(g_1,\ldots,g_m)$ and denote
the class of these sequences by $\Fclass_T^m$. We retain the value $c$ for
$t\leq r_0$ at every iteration and define the candidate paths by
\[
 \volhat{t,0}(c)=c,\qquad
 \volhat{t,r}(c,g_{1:r})
 =g_r\{Y_{t-1:t-p},\volhat{t-1:t-q,r-1}(c,g_{1:r-1})\}.
\]
Within this argument, $X_{t,m-1}$ denotes the regressor generated by
$c,g_{1:m-1}$. We define $e_{t,m}=\volsq_t-g_m(X_{t,m-1})$ and
$e_{S,t,m}=\volsq_t-f_{\snet}(X_{t,m-1})$. These errors refer to deterministic
candidates.

For each candidate sequence, $e_{S,t,m}-e_{t,m}$ depends only on returns
observed before time $t$. Its absolute value is at most $2\Bmax$.
Since $\mds_t=\volsq_t(\varepsilon_t^2-1)$,
Assumption~\ref{ass:innov-up} gives
$\E\{\exp(\lambda\mds_t)\mid\filt{t-1}\}\leq\exp(C\lambda^2)$ for small
$|\lambda|$. Consequently,
\[
 \E\exp\left\{\lambda\sum_t\mds_t(e_{S,t,m}-e_{t,m})
 -C\lambda^2\sum_t(e_{S,t,m}-e_{t,m})^2\right\}\leq1.
\]
For sufficiently small $\eta>0$, Markov's inequality with $\lambda$
proportional to $\eta$ gives, for every $x>0$,
\[
 \Prob\left[
 2P_T\{\mds_t(e_{S,t,m}-e_{t,m})\}
 >\eta P_T(e_{S,t,m}-e_{t,m})^2+\frac{C_\eta x}{n_T}
 \right]\leq e^{-x}.
\]

Next, we prove a uniform version of the preceding bound by covering the
initial-value interval and the class of network sequences.
For $g_{1:M},\tilde g_{1:M}\in\Fclass_T^M$, we define
\[
 \|g_{1:M}-\tilde g_{1:M}\|_{\max,\infty}
 :=\max_{1\leq r\leq M}\sup_{x\in\mathcal X_T}
 |g_r(x)-\tilde g_r(x)|.
\]
We set $\delta=(1-B')T^{-3}$. We denote the covering number of
$\Fclass_T^M$ at radius $\delta$ under the norm $\|\cdot\|_{\max,\infty}$
by $\mathcal N(\delta,\Fclass_T^M,\|\cdot\|_{\max,\infty})$, which is the
minimum number of representative sequences in $\Fclass_T^M$ needed to
approximate every sequence within distance $\delta$.

To control the covering error in the generated variance paths, we compare
$(c,g_{1:M})$ and $(\tilde c,\tilde g_{1:M})$ with
$|c-\tilde c|\leq\delta$ and
$\|g_{1:M}-\tilde g_{1:M}\|_{\max,\infty}\leq\delta$.
For $(y,s),(y,\tilde s)\in\mathcal X_T$, the triangle inequality and the
variance-input Lipschitz bound give
\[
 |g_r(y,s)-\tilde g_r(y,\tilde s)|
 \leq\|g_r-\tilde g_r\|_{\infty,\mathcal X_T}
      +B'\|s-\tilde s\|_\infty
 \leq\delta+B'\|s-\tilde s\|_\infty.
\]
We denote the largest absolute difference between the two stage-$r$
variance paths over $1\leq t\leq T$ by $u_r$. The initial and retained
boundary values differ by at most $\delta$. In particular, $u_0\leq\delta$.
On the return-truncation event, the two paths use the same return inputs.
The displayed bound therefore gives
\[
 u_r\leq\delta+B'u_{r-1}
 \leq\delta\sum_{j=0}^{r-1}(B')^j+(B')^ru_0
 \leq\frac{\delta}{1-B'}=T^{-3}.
\]

We approximate the initial value by a grid of at most $C/\delta$ points
in $[\Bmin,\Bmax]$. Pairing these points with a $\delta$-cover of
$\Fclass_T^M$ gives at most
$(C/\delta)\mathcal N(\delta,\Fclass_T^M,\|\cdot\|_{\max,\infty})$
candidate pairs. The same cover works at every iteration because the
distance controls each of the $M$ networks.
We write $\tilde X_{t,m-1}$ for the regressor generated by the representative
candidate. Since $e_{S,t,m}-e_{t,m}=(g_m-f_{\snet})(X_{t,m-1})$, the
approximation error in this difference satisfies
\[
 \bigl|(g_m-f_{\snet})(X_{t,m-1})
       -(\tilde g_m-f_{\snet})(\tilde X_{t,m-1})\bigr|
 \leq\delta+2B'u_{m-1}\leq CT^{-3}.
\]
Here $\delta$ is the covering error for $g_m$. The term $2B'u_{m-1}$
bounds the effect of changing the variance inputs of $g_m$ and the comparison
network $f_{\snet}$.

We apply the candidate inequality to every representative pair and every
$m\leq M$. With
$x=\log\mathcal N(\delta,\Fclass_T^M,\|\cdot\|_{\max,\infty})
+\log(CM/\delta)+A\log T$, the union bound gives
\[
\begin{aligned}
 &\Prob\{\text{at least one of these candidate inequalities fails}\}\\
 &\quad\leq\sum_{\text{representative pairs}}\sum_{m=1}^M e^{-x}
 \leq\frac{CM}{\delta}
 \mathcal N(\delta,\Fclass_T^M,\|\cdot\|_{\max,\infty})e^{-x}
 =T^{-A}.
\end{aligned}
\]
For sufficiently large $C_Y$ and $C_A$, the tail bounds give
\[
 \Prob\!\left(\max_t|Y_t|>C_Y\sqrt{\log T}\right)
 +\Prob\!\left(\max_t|\mds_t|>C_A\log T\right)\leq CT^{-A}.
\]
Outside these exceptional events, the preceding approximation bound applies
to every candidate sequence. Bounded network outputs also give
\[
 P_T\{(\tilde g_m-f_{\snet})(\tilde X_{t,m-1})\}^2
 \leq P_T(e_{S,t,m}-e_{t,m})^2+CT^{-3}.
\]
We can therefore transfer the candidate inequality from the representative
to the original sequence. Simultaneously for all candidates and $m\leq M$,
\begin{align*}
 &2P_T\{\mds_t(e_{S,t,m}-e_{t,m})\}\\
 &\quad\leq2P_T\{\mds_t(\tilde g_m-f_{\snet})(\tilde X_{t,m-1})\}
       +CT^{-3}\max_t|\mds_t|\\
 &\quad\leq\eta P_T\{(\tilde g_m-f_{\snet})(\tilde X_{t,m-1})\}^2
       +\frac{C_\eta x}{n_T}+CT^{-3}\max_t|\mds_t|\\
 &\quad\leq\eta P_T(e_{S,t,m}-e_{t,m})^2
       +\frac{C_\eta x}{n_T}+C_\eta T^{-3}(1+\log T)\\
 &\quad\leq\eta P_T(e_{S,t,m}-e_{t,m})^2+\frac{C_\eta}{n_T}
 \log\mathcal N(\delta,\Fclass_T^M,\|\cdot\|_{\max,\infty})
 \\
 &\qquad\quad+\frac{C_\eta}{n_T}\{\log(CM/\delta)+A\log T\}
 +C_\eta T^{-2},
\end{align*}
with probability at least $1-CT^{-A}$ by the two probability bounds above.

A product cover and Proposition~\ref{prop:network-up} give
\[
\begin{aligned}
\log\mathcal N(\delta,\Fclass_T^M,\|\cdot\|_{\max,\infty})
&\leq M\log\mathcal N(\delta,\Fclass_T,
\|\cdot\|_{\infty,\mathcal X_T})\\
&\leq CM\snet\dnet\log(CT/\delta).
\end{aligned}
\]
Since $\delta=(1-B')T^{-3}$, substitution gives the stochastic bound
\begingroup\small
\[
 2P_T\{\mds_t(e_{S,t,m}-e_{t,m})\}
 \leq\eta P_T(e_{S,t,m}-e_{t,m})^2
 +C_\eta\frac{M\snet\dnet\log T+A\log T}{n_T}+C_\eta T^{-2}.
\]
\endgroup
This bound holds uniformly over all candidate sequences and $m\leq M$
with probability at least $1-CT^{-A}$.

\smallskip
\noindent\textbf{Error recursion and convergence rate.}\par\nobreak
We now combine the stochastic bound with the initial error decomposition
to obtain a recursion for $\Delta_{T,m}^{\rm in}$.
The uniform bound applies to $c=\widehat c$ and $g_r=\widehat f_r$.
The errors $e_{t,m}$ and $e_{S,t,m}$ below are evaluated along this estimated
sequence.

To take expectations, we use $2A$ in place of $A$ in the probability bound.
The failure probability is then at most $CT^{-2A}$. Since the errors are bounded
and $\E(P_T|\mds_t|)^2\leq\E P_T\mds_t^2\leq C$, Cauchy--Schwarz bounds
the contribution of the exceptional event to the expected stochastic term by $CT^{-A}$.
Using $(e_{S,t,m}-e_{t,m})^2\leq2e_{t,m}^2+2e_{S,t,m}^2$, we obtain
\begin{align*}
 2\E P_T\{\mds_t(e_{S,t,m}-e_{t,m})\}
 &\leq2\eta(\Delta_{T,m}^{\rm in}+\E P_Te_{S,t,m}^2)\\
 &\quad+C_\eta\left\{\frac{M\snet\dnet\log T+A\log T}{n_T}
       +T^{-2}+T^{-A}\right\}.
\end{align*}
\begingroup\displaywidowpenalty=10000
We substitute this expectation bound into the initial error decomposition and use
the earlier bound for $\E P_Te_{S,t,m}^2$. Moving
$2\eta\Delta_{T,m}^{\rm in}$ to the left and dividing by $1-2\eta$ gives
the coefficient $(1+2\eta)(1+\tau)\Dcon^2/(1-2\eta)$ on
$\Delta_{T,m-1}^{\rm in}$. Since $\Dcon<1$, we set
$\rho=(1+\Dcon^2)/2<1$ and choose $\tau>0$ and $\eta\in(0,1/2)$
sufficiently small that this coefficient is at most $\rho$.
Absorbing $T^{-2}$ into the sampling term gives
\begingroup\small
\[
 \Delta_{T,m}^{\rm in}\leq\rho\Delta_{T,m-1}^{\rm in}
 +C\left\{\|f_{\snet}-\ftrue\|_{\infty,\mathcal X_T}^2
 +\frac{M\snet\dnet\log T+A\log T+\qstar}{n_T}+T^{-A}\right\}.
\]
\endgroup
\endgroup
Since $\Delta_{T,0}^{\rm in}\leq(\Bmax-\Bmin)^2$, iteration gives
\begingroup\small
\[
 \Delta_{T,M}^{\rm in}\leq C\left[\rho^M+\frac{1}{1-\rho}
 \left\{\|f_{\snet}-\ftrue\|_{\infty,\mathcal X_T}^2
 +\frac{M\snet\dnet\log T+A\log T+\qstar}{n_T}+T^{-A}\right\}\right].
\]
\endgroup
For $\qstar=0$, the preceding-iteration and initialization terms are absent.

Finally, we balance approximation and sampling errors.
Since $M,\dnet=O(\log T)$ and $n_T\asymp T$, the sampling term is at most
$C\snet(\log T)^3/T$. The approximation term is bounded by
$C(\log T)^c\snet^{-2\hol/(\fdim)}$.
The theorem's choice $\snet\asymp T^{(\fdim)/(\fdim+2\hol)}$ balances
these terms up to logarithmic factors. We choose the constant in
$M\asymp\log T$ sufficiently large that the initialization term $\rho^M$
is no larger than $T^{-2\hol/(\fdim+2\hol)}$.
The boundary and tail remainders are smaller because the lag orders do not
depend on $T$ and $A>2$. This proves Theorem~\ref{thm:main-upfront}.
\end{proof}

\section{Proof of Theorem~\ref{thm:lower}}\label{app:lower}

We aim to show that every estimator $\widetilde{\boldsymbol v}$ based on
$Y_1,\ldots,Y_T$ satisfies the following lower bound for sufficiently large $T$,
\[
 \sup_{\ftrue\in\Cclass}
 \frac1{T-r_*}\sum_{t=r_*+1}^{T}
 \E_{\ftrue}(\widetilde v_t-\volsq_t)^2
 \geq c_{\rm LB}T^{-2\hol/(2\hol+\fdim)},
\]
where $r_*=\max(\pstar,\qstar)$. For this purpose, we will construct
stationary alternatives with local perturbations that are difficult to
distinguish. We introduce an observable reference path to handle the latent
variance inputs. We then use a Bayes-risk argument to combine the frequency
of visits to each perturbation region with uncertainty about its sign.
Finally, we bound these quantities jointly to obtain the required lower bound.

\begin{proof}
\noindent\textbf{Construction of alternative recursion maps.}
We construct a subfamily of $\Cclass$ whose members are difficult to
distinguish using the observed returns. We obtain this family by adding local
perturbations to a smooth reference map. To keep the perturbed functions in
$\Cclass$, we choose the reference map with strict margins in the variance,
H\"older, and contraction bounds. For
$x=(y,s)\in\R^{\pstar}\times\R^{\qstar}$, we set
$\fbase(x)=c+\alpha\tanh(y_{\pstar})$ and choose $c$ and a sufficiently small
$\alpha>0$ such that
$\Bmin<\inf_x\fbase(x)\leq\sup_x\fbase(x)<\Bmax$ and
$\|\fbase\|_{\mathcal H^\hol}<\Bmax$.
These margins do not depend on $T$. Dependence on the last return allows the
stationary state to have a density in all $\fdim$ coordinates.

To localize the perturbations, we choose compact rectangles
$\mathcal Q\Subset\mathcal Q_0$ inside
$\R^{\pstar}\times(c-\alpha,c+\alpha)^{\qstar}$.  When $\qstar=0$, the
second factor is absent. The perturbations lie in $\mathcal Q$.
We bound the stationary density on $\mathcal Q_0$.
Let $\mbump\in C_c^\infty((0,1)^{\fdim})$ be nonnegative, positive on an
inner cube, and nonconstant in every coordinate. We rescale $\mbump$ to satisfy
$0\leq\mbump\leq1$ and $\|\mbump\|_{\mathcal H^\hol}\leq1$.
For $\hcube\in(0,1)$, we place $\Ncell\asymp\hcube^{-(\fdim)}$ disjoint cubes of
side $\hcube$ inside $\mathcal Q$. Each bump's support stays a constant
fraction of $\hcube$ from its cube's boundary. If $z_j$ is the
lower corner of cube $j$, we set
$\bumpj{j}(x)=\mbump\{(x-z_j)/\hcube\}$ and
$\amp=\ampz\hcube^\hol$, where $\ampz>0$ is a sufficiently small constant
independent of $T$.
For $\theta\in\{-1,1\}^{\Ncell}$, we define
\begin{equation}
 B_\theta(x)=\sum_{j=1}^{\Ncell}\theta_j\bumpj{j}(x),
 \qquad
 \fpert(x)=\fbase(x)+\amp B_\theta(x).
 \label{eq:lb-family}
\end{equation}

We next verify that each perturbed map remains in $\Cclass$.
Disjoint supports and H\"older scaling give
$\|\amp B_\theta\|_{\mathcal H^\hol}
\leq C\amp\hcube^{-\hol}=C\ampz$ and
$\|B_\theta\|_{\rm Lip}\leq C\hcube^{-1}$.
If $\qstar\geq1$, we use the weights $\omega_j=1/\qstar$ in
Assumption~\ref{ass:lip-up}.  Since $\fbase$ does not depend on $s$,
$|\fpert(y,s)-\fpert(y,s')|
\leq C\ampz\hcube^{\hol-1}\|s-s'\|_\omega$.
Because $\hol\geq1$, we choose $\ampz$ small enough to preserve the variance
and H\"older bounds and keep the variance-input Lipschitz coefficient at most
$\bar D$. Contraction is
automatic when $\qstar=0$. Each bump varies in every coordinate, ensuring
that every $\fpert$ has structural orders exactly $(\pstar,\qstar)$.

\noindent\textbf{Stationary density on the perturbation region.}
We next establish stationarity and density bounds on $\mathcal Q_0$ to
control regional visit frequencies. Each alternative has Markov state
$X_t^\theta=(Y_{t-1},\ldots,Y_{t-\pstar},
\sigma_{t-1}^2,\ldots,\sigma_{t-\qstar}^2)$. To derive its transition
density, we express the state after $\fdim$ updates from $x$ as
$H_{\theta,x}(u)$, where $u=(u_1,\ldots,u_{\fdim})$ collects the consecutive
returns. Under $\fbase$, this
map is independent of $x$ and, up to coordinate order, equals
\[
 H_0(u)=
 \bigl(u_{\fdim},\ldots,u_{\qstar+1},
 c+\alpha\tanh(u_{\qstar}),\ldots,
 c+\alpha\tanh(u_1)\bigr).
\]
The variance coordinates are absent when $\qstar=0$. The map $H_0$ is one-to-one onto
$\R^{\pstar}\times(c-\alpha,c+\alpha)^{\qstar}$. We choose a compact rectangle
$\mathcal U$ whose interior contains the preimage under $H_0$ of a
neighborhood of $\mathcal Q_0$.  Thus $H_0(u)$ stays a positive distance from
$\mathcal Q_0$ for $u\notin\mathcal U$.  On $\mathcal U$, $H_0$ is
bi-Lipschitz.  Its absolute Jacobian determinant
$\alpha^{\qstar}\prod_{j=1}^{\qstar}\operatorname{sech}^2(u_j)$ is bounded away
from zero.

We compare $H_{\theta,x}$ with $H_0$ to justify the change of variables.
The uniform variance-perturbation bound $\amp$, the Lipschitz bound on
$B_\theta$, and the chain rule over $\fdim$ updates give
\[
 \sup_{x,u}\|H_{\theta,x}(u)-H_0(u)\|\leq C\amp,
 \qquad
 \sup_x\sup_{u\in\mathcal U}
 \|\nabla H_{\theta,x}(u)-\nabla H_0(u)\|
 \leq C\amp\hcube^{-1}.
\]
For sufficiently small $\ampz$, the derivative bound and the bi-Lipschitz
property of $H_0$ make $H_{\theta,x}$ one-to-one on $\mathcal U$, with a
uniformly bounded inverse Jacobian. The first bound keeps $H_{\theta,x}(u)$
outside $\mathcal Q_0$ whenever $u\notin\mathcal U$. A degree argument using
the closeness to $H_0$ on the boundary of $\mathcal U$ shows that
$H_{\theta,x}(\mathcal U)$ contains $\mathcal Q_0$.
Conditional on $x$, the joint density of $u$ is a product of conditional
Gaussian densities. Since each variance lies in $[\Bmin,\Bmax]$, this density
is uniformly bounded above and away from zero on $\mathcal U$. We write
$k_\theta^{\fdim}(x,z)$ for the $\fdim$-step transition density of the
state process.  A change of variables gives
$0<c\leq k_\theta^{\fdim}(x,z)\leq C<\infty$ uniformly for
$x\in\R^{\pstar}\times[\Bmin,\Bmax]^{\qstar}$ and $z\in\mathcal Q_0$.
The density lower bound gives a common Doeblin minorization on
$\mathcal Q_0$. Each state process therefore has a unique stationary
distribution and is uniformly geometrically ergodic. Integrating the
transition-density bounds against this distribution gives a density
$\pi_\theta$ on $\mathcal Q_0$ with $0<c\leq\pi_\theta(z)\leq C$.
We write $\E_\theta$ for expectation under this stationary model.
The stationary state is geometrically absolutely regular, uniformly over $\theta$
\citep[Ch.~16]{meyntweedie2009,bradley2005}.

\noindent\textbf{Reduction to posterior uncertainty.}
We assign independent uniform priors to the signs to express risk through
posterior uncertainty. We denote mixture expectation by $\Emix$.
We also allow the estimator to observe
$\mathcal O_T=\sigma(X_1^\theta,Y_1,\ldots,Y_T)$, including the initial
state. This additional information can only reduce the minimum risk.
The worst-case risk dominates the mixture-average
risk. The posterior mean minimizes average squared loss. With $n_*=T-r_*$,
we therefore have
\[
 \sup_{\ftrue\in\Cclass}\frac1{n_*}\sum_{t=r_*+1}^{T}
 \E_{\ftrue}(\widetilde v_t-\volsq_t)^2
 \geq\frac1{n_*}\sum_{t=r_*+1}^{T}
 \Emix\{\Var(\sigma_t^{2,\theta}\mid\mathcal O_T)\}.
\]

The latent variance coordinates of $X_t^\theta$ depend on the unknown signs.
We replace them by observable reference values to bound the posterior
variance. We define the reference variance
$\bar v_t=\fbase(X_t^\theta)=c+\alpha\tanh(Y_{t-\pstar})$.
The corresponding regressor is
$\bar X_t=(Y_{t-1},\ldots,Y_{t-\pstar},
\bar v_{t-1},\ldots,\bar v_{t-\qstar})$.
For $t=r_*+1,\ldots,T$, the presample returns in $X_1^\theta$ allow us to
construct $\bar X_t$ from $\mathcal O_T$. The construction gives
$\sigma_t^{2,\theta}-\bar v_t=\amp B_\theta(X_t^\theta)$.  Since the bump
supports are disjoint, $|B_\theta|\leq1$. The latent and reference variance
coordinates therefore differ by at most $\amp$. The Lipschitz bound on
$B_\theta$ then gives
\begin{equation}
 \|X_t^\theta-\bar X_t\|_2\leq C\amp,
 \qquad
 \left|\amp B_\theta(X_t^\theta)
       -\amp B_\theta(\bar X_t)\right|
 \leq C\frac{\amp^2}{\hcube}.
 \label{eq:lb-reference-error}
\end{equation}

We measure the weighted visit frequency by
$\desfreq{j}=n_*^{-1}\sum_{t=r_*+1}^{T}\bumpj{j}(\bar X_t)^2$.
The posterior variance $\postvar{j}=\Var(\theta_j\mid\mathcal O_T)$ measures
the remaining uncertainty about sign $j$. Subtracting the observable
$\bar v_t$ leaves the conditional
variance unchanged. At most one bump is nonzero at $\bar X_t$. The
conditional-variance inequality and the reference approximation give
\[
 \Var(\sigma_t^{2,\theta}\mid\mathcal O_T)
 \geq\frac{\amp^2}{2}\sum_{j=1}^{\Ncell}
 \bumpj{j}(\bar X_t)^2\postvar{j}
 -C\frac{\amp^4}{\hcube^2}.
\]
It therefore suffices to establish the joint lower bound
$\Emix\{\desfreq{j}\postvar{j}\}\geq c\hcube^{\fdim}$ for every $j$.
Summing over the $\Ncell\asymp\hcube^{-(\fdim)}$ regions then gives a lower
bound of order $\amp^2$, apart from the remainder $C\amp^4/\hcube^2$.

\noindent\textbf{Frequency and uncertainty of local perturbations.}
We choose $\hcube=T^{-1/(2\hol+\fdim)}$ to balance perturbation size
against sample information. For sufficiently small $\ampz$, the frequency
moments satisfy, uniformly in $j$,
$\Emix\desfreq{j}\asymp\hcube^{\fdim}$ and
$\Emix(\desfreq{j}^2)\leq C\hcube^{2(\fdim)}$.
Indeed, the reference approximation and $\amp\leq\ampz\hcube$ show that
$\bar X_t$ differs from the stationary state $X_t^\theta$ by less than a
small fraction of $\hcube$. The density bounds on $\mathcal Q_0$
and the inner region on which $\mbump$ is positive then give
$c\hcube^{\fdim}\leq\E_\theta\{\bumpj{j}(\bar X_t)^2\}
\leq C\hcube^{\fdim}$.  The bounded sequence
$Z_{t,j}=\bumpj{j}(\bar X_t)^2$ is a function of
$(X_{t-\qstar}^\theta,\ldots,X_t^\theta)$ and inherits geometric absolute
regularity. Boundedness and $\E_\theta Z_{t,j}\leq C\hcube^{\fdim}$ control
short-lag covariances. Geometric absolute regularity controls long-lag
covariances. Combining these bounds gives, for some $\lambda\in(0,1)$,
\[
 \begin{aligned}
 |\Cov_\theta(Z_{t,j},Z_{t+k,j})|
 &\leq C\min\{\hcube^{\fdim},\lambda^{(k-\qstar)_+}\},\\
 \Var_\theta(\desfreq{j})
 &\leq \frac{C\hcube^{\fdim}\log(1/\hcube)}{T}.
 \end{aligned}
\]
Since $T\hcube^{\fdim}/\log(1/\hcube)\to\infty$, the stated moment bounds follow.

To control the information about each sign, we condition on $\theta_{-j}$
and compare the models with
$\theta_j=1$ and $\theta_j=-1$. Write $P_a$ and $\E_a$ for probability
and expectation under sign $a\in\{+,-\}$.
To compare likelihoods, we generate variance paths $v_t^\pm$ and regressors
$X_t^\pm$ from the same initial state and return sequence.
With $e_t=v_t^+-v_t^-$, contraction gives
\begin{align*}
 |e_t|&\leq\bar D
 \left\{\sum_{\ell=1}^{\qstar}\omega_\ell e_{t-\ell}^2\right\}^{1/2}
 +2\amp|\bumpj{j}(X_t^-)|,\\
 \sum_{t=1}^{T}e_t^2
 &\leq C\amp^2\sum_{t=1}^{T}\bumpj{j}(X_t^-)^2,
 &\max_t|e_t|&\leq C\amp.
\end{align*}
The contraction term is absent when $\qstar=0$. We square the first
inequality and apply Young's inequality with a lagged-error coefficient
between $\bar D^2$ and one. Summing over dates and absorbing lagged errors
gives the squared-path bound. The density bound applies directly to $X_t^-$
under the minus model. Under the plus model, $X_t^-$ differs from $X_t^+$ by
at most $C\amp\leq C\hcube$. We therefore apply the density bound to
$X_t^+$ on an $O(\hcube)$ enlargement of the bump support. Under either sign,
$\E_\pm\sum_{t=1}^{T}e_t^2\leq
CT\amp^2\hcube^{\fdim}$.

We use this variance-path bound to control the log-likelihood ratio.
Let $q_v$ denote the $N(0,v)$ density and $\logodds{j}$ the transition
log-likelihood ratio of the plus model against the minus model, conditional
on the initial state. For $Y\sim N(0,v)$ and $v,w\in[\Bmin,\Bmax]$,
Gaussian moments give
\[
 \E_v[\{\log(q_v/q_w)(Y)\}^2]\leq C(v-w)^2,\qquad
 \KL\{N(0,v),N(0,w)\}\leq C(v-w)^2.
\]
We decompose $\logodds{j}$ into a martingale sum and a sum of conditional
means. By the Gaussian and variance-path bounds, the martingale sum has
second moment at most $CT\amp^2\hcube^{\fdim}=C\ampz^2$.
The conditional-mean sum has expected absolute value at most $C\ampz^2$.
Both bounds hold under either sign. Markov's inequality therefore gives
\[
 \sup_{\theta_{-j}}\max_{a\in\{+,-\}}
 P_a\{|\logodds{j}|>1\}\leq C\ampz^2.
\]

Since the estimator also observes the initial state, we include its
stationary density in the posterior odds.
Let $A_0=\{X_1^\theta\in\mathcal Q_0\}$. On $A_0$, the ratio of the
stationary densities $\pi_+$ and $\pi_-$ is bounded above and away from zero.
Bayes' formula gives
\[
 \frac{\Prob(\theta_j=1\mid\mathcal O_T,\theta_{-j})}
      {\Prob(\theta_j=-1\mid\mathcal O_T,\theta_{-j})}
 =\frac{\pi_+(X_1^\theta)}{\pi_-(X_1^\theta)}\exp(\logodds{j}).
\]
On $A_0\cap\{|\logodds{j}|\leq1\}$, the conditional variance of
$\theta_j$ given $(\mathcal O_T,\theta_{-j})$ is therefore bounded below by a
positive constant. We combine this bound with regional visits after omitting
$O\{\log(1/\hcube)\}$ initial dates. Geometric absolute regularity controls
their dependence on the initial state. The density lower bound then gives
$\Emix\{\desfreq{j}\mathbf 1(A_0)\}\geq c\hcube^{\fdim}$.
The omitted fraction is negligible relative to $\hcube^{\fdim}$ because
$T\hcube^{\fdim}/\log(1/\hcube)\to\infty$. We use the law of total variance
to average over $\theta_{-j}$. Cauchy--Schwarz, the second moment of
$\desfreq{j}$, and the probability bound for $|\logodds{j}|>1$ give
\begin{align*}
 \Emix\{\desfreq{j}\postvar{j}\}
 &\geq c\Emix\left[\desfreq{j}\mathbf 1(A_0)
 \mathbf 1\{|\logodds{j}|\leq1\}\right]\\
 &\geq c\hcube^{\fdim}-C\ampz\hcube^{\fdim}.
\end{align*}
Choosing $\ampz$ sufficiently small proves the required joint bound on
visits and posterior uncertainty.

\noindent\textbf{Completion of the lower bound.}
We substitute the joint bound into the posterior-risk decomposition to obtain
\begin{align*}
 \frac1{n_*}\sum_{t=r_*+1}^{T}
 \Emix\{\Var(\sigma_t^{2,\theta}\mid\mathcal O_T)\}
 &\geq c\amp^2\sum_{j=1}^{\Ncell}
 \Emix\{\desfreq{j}\postvar{j}\}
 -C\frac{\amp^4}{\hcube^2}\\
 &\geq c\amp^2-C\frac{\amp^4}{\hcube^2}.
\end{align*}

Here $\Ncell\hcube^{\fdim}\asymp1$. The remainder is $o(\amp^2)$ when
$\hol>1$. When $\hol=1$, we absorb it
into the leading term by reducing $\ampz$.  The selected bandwidth gives
$\amp^2=\ampz^2\hcube^{2\hol}
\asymp T^{-2\hol/(2\hol+\fdim)}$.
The Bayes-risk reduction therefore proves the required minimax lower bound.
\end{proof}

\section{From in-sample risk to population and forecast risk}
\label{app:population-transfer}

Under Assumption~\ref{ass:dep-up}, we extend the in-sample analysis to
recursion-map estimation, future forecasts, and expected excess QLIKE loss.
Let $(Y_t',(\sigma_t')^2)_{t\in\Z}$ be a stationary copy independent of the
estimation sample. We write $X_{t,*}'$ for its regressor vector formed from
the $p$ return lags and $q$ true variance lags.

\begin{lemma}[Population and future-forecast risk]
\label{lem:population-transfer}
Suppose the conditions of Theorem~\ref{thm:main-upfront} hold and
Assumption~\ref{ass:dep-up} is satisfied.  There are constants $C>0$, $c\geq0$,
and $\lambda_{\mathrm F}\in(0,1)$ such that the following holds.  Write
$r_T=T^{-2\hol/(\fdim+2\hol)}(\log T)^c$.  For every sufficiently large $T$
and every $k\geq1$,
\begin{align}
 \E\{\widehat f_M(X_{0,*}')-\ftrue(X_{0,*}')\}^2
 &\leq Cr_T, \notag\\
 \E\{\widehat\sigma^2_{T+k\mid T}-\volsq_{T+k}\}^2
 &\leq Cr_T+C\lambda_{\mathrm F}^{k}.
 \label{eq:apt-population-bounds}
\end{align}
There are constants $0<c_{\mathrm Q}<C_{\mathrm Q}<\infty$, depending only on
$\Bmin$ and $\Bmax$, such that expected excess QLIKE at date $T+k$ lies
between $c_{\mathrm Q}$ and $C_{\mathrm Q}$ times the second mean-squared
error in \eqref{eq:apt-population-bounds}.
\end{lemma}

\begin{proof}
We first separate stationary risk into approximation, propagated variance
error, and stochastic terms. We reuse the in-sample bound for
squared-return noise. We bound the additional sample-to-population
deviations under mixing and obtain a risk recursion. This yields the
recursion-map bound, which we then extend to forecasts and QLIKE loss.

\smallskip
\noindent\textbf{From sample loss to stationary risk.}\par\nobreak
We begin with a risk decomposition and then control its stochastic terms
uniformly over the candidate network sequences.
For $m\leq M$, we consider a deterministic initial value
$b\in[\Bmin,\Bmax]$ and deterministic networks
$g_1,\ldots,g_{m-1}\in\Fclass_T$.
We write \mbox{$\mathcal A=(b,g_{1:m-1})$}.
The preceding networks generate the regressor
\[
X_{t,T}^{\mathcal A}
=\{Y_{t-1:t-p},\volhat{t-1:t-q,m-1}(b,g_{1:m-1})\}.
\]
For a current candidate $f\in\Fclass_T$, set
$L_{t,T}^{\mathcal A,f}
=\{\volsq_t-f(X_{t,T}^{\mathcal A})\}^2$.
We use the stationary copy $(Y_t',(\sigma_t')^2)_{t\in\Z}$ to remove
the sample boundary. This copy is independent of $(Y_1,\ldots,Y_T)$.
For every $t\in\Z$, we set
$\widehat\sigma_{t,0}^{2\prime}(\mathcal A)=b$. The updates for
$r=1,\ldots,m-1$ and the regressor at iteration $m$ are
\begin{align*}
 \widehat\sigma_{t,r}^{2\prime}(\mathcal A)
 &=g_r\{Y_{t-1:t-p}',
        \widehat\sigma_{t-1:t-q,r-1}^{2\prime}(\mathcal A)\},\\
 X_t^{\mathcal A}
 &=\{Y_{t-1:t-p}',\widehat\sigma_{t-1:t-q,m-1}^{2\prime}(\mathcal A)\}.
\end{align*}
We define the corresponding loss and stationary risk by
$L_t^{\mathcal A,f}=\{(\sigma_t')^2-f(X_t^{\mathcal A})\}^2$ and
$Q_{\mathcal A}(f)=\E' L_0^{\mathcal A,f}$, where $\E'$ averages over the
independent copy. For deterministic $\mathcal A$ and $f$, these losses form
a stationary process.
Suppose $f$ has no larger empirical squared-return loss than the comparison
network $f_{\snet}$ at $X_{t,T}^{\mathcal A}$. As in the in-sample proof,
we expand this comparison using $Y_t^2=\volsq_t+\mds_t$. Adding and
subtracting the sample conditional-variance losses gives
\begin{align*}
 Q_{\mathcal A}(f)
 \leq{}&Q_{\mathcal A}(f_{\snet})
 +2P_T\left[\mds_t
 \{f(X_{t,T}^{\mathcal A})-f_{\snet}(X_{t,T}^{\mathcal A})\}\right]\\
 &+\{Q_{\mathcal A}(f)-P_TL_{t,T}^{\mathcal A,f}\}
 +\{P_TL_{t,T}^{\mathcal A,f_{\snet}}-Q_{\mathcal A}(f_{\snet})\}.
\end{align*}
The comparison-network risk contains both approximation error and error
from the generated variance inputs. The second term is squared-return
noise. The last two terms compare sample losses with stationary risks.
We first separate the two sources of comparison-network error. Adding and
subtracting $\ftrue(X_0^{\mathcal A})$, then applying Young's inequality
and contraction, controls the error from the generated variance inputs.
Stationarity gives the same mean-squared variance error at every lagged date. For
$\tau>0$ and any $A>2$, the return-tail bound therefore gives
\[
 Q_{\mathcal A}(f_{\snet})
 \leq(1+\tau)\Dcon^2
 \E'|\widehat\sigma_{0,m-1}^{2\prime}(\mathcal A)-(\sigma_0')^2|^2
 +C_\tau\|f_{\snet}-\ftrue\|_{\infty,\mathcal X_T}^2+CT^{-A}.
\]
The preceding-iteration term is absent when $\qstar=0$. We retain the
approximation error in this form until the final rate calculation.
The uniform stochastic bound from the proof of
Theorem~\ref{thm:main-upfront} already controls the squared-return noise.
For sufficiently small $\eta>0$, it gives the following sample-loss
comparison with probability at least $1-CT^{-A}$,
\begingroup\small
\begin{equation}
 (1-2\eta)P_TL_{t,T}^{\mathcal A,f}
 \leq(1+2\eta)P_TL_{t,T}^{\mathcal A,f_{\snet}}
 +C_\eta\frac{M\snet\dnet\log T+A\log T}{n_T}+C_\eta T^{-2}.
 \label{eq:apt-empirical-oracle}
\end{equation}
\endgroup
This bound holds simultaneously over $m\leq M$, deterministic
$\mathcal A$, and candidates $f$ satisfying the empirical loss comparison.

It remains to control the sample-to-population deviations in the
decomposition. To use the mixing condition on returns, we first express
the losses in terms of a finite return segment.
The network inputs use at most $R_M=p+(M-1)q$ preceding returns, but the
true variance may depend on the entire past. We therefore approximate the
true variance by a truncated filter. When $\qstar\geq1$, we initialize the
variance state at date $t-(M-1)q$ at $(\Bmin,\ldots,\Bmin)$ and apply
$\ftrue$ recursively. The resulting
$\sigma_t^{2,\mathrm{tr}}$ depends only on
$Y_{t-R_M},\ldots,Y_{t-1}$. When $\qstar=0$, take
$\sigma_t^{2,\mathrm{tr}}=\volsq_t$. With $\rho=(1+\Dcon^2)/2$,
contraction gives
\[
 |\volsq_t-\sigma_t^{2,\mathrm{tr}}|^2\leq C\delta_M^{\rm tr},
 \qquad
 \delta_M^{\rm tr}=
 \begin{cases}
  0, & \qstar=0,\\
  \Dcon^{2\lfloor(M-1)q/\qstar\rfloor}\leq2\rho^M, & \qstar\geq1.
 \end{cases}
\]
Young's inequality bounds the effect of this replacement on either loss by
a factor $1+\eta$ and an additive $C_\eta\delta_M^{\rm tr}$.

We compare truncated losses using blocks of length
$\ell_{T,M}=R_M+c_{\rm b}\log T$, where $c_{\rm b}>0$ is sufficiently
large. We collect the sampling and tail terms in
\[
 \mathfrak E_{T,M}
 =C\frac{(1+\log T)\ell_{T,M}}{n_T}
       \{M\snet\dnet\log T+A\log T\}+CT^{-A}.
\]
With probability at least $1-CT^{-A}$, we establish the following comparison uniformly over
iterations, deterministic $\mathcal A$, and candidate networks,
\begin{equation}
 \left|P_TL_{t,T}^{\mathcal A,f}-Q_{\mathcal A}(f)\right|
 \leq\eta\{P_TL_{t,T}^{\mathcal A,f}+Q_{\mathcal A}(f)\}
 +C_\eta\{\mathfrak E_{T,M}+\delta_M^{\rm tr}\}.
 \label{eq:apt-stationary-transfer}
\end{equation}

To prove this comparison, we omit the first $R_M$ eligible observations to remove the effect of the
sample boundary on the generated inputs.
After the variance replacement, stationarity gives the same loss distribution
on the observed and independent paths. The return-tail bound allows us to
restrict the covering argument to $\mathcal X_T$. As in the proof of
Theorem~\ref{thm:main-upfront}, a $T^{-3}$-cover of initial values and network
sequences induces a bounded-loss cover with log cardinality at most
$CM\snet\dnet\log T$.

We divide the remaining dates into blocks of length $\ell_{T,M}$ and
extend each block backward by $R_M$ dates. The extended return segments
from odd blocks are separated by $c_{\rm b}\log T$ dates, as are those from
even blocks. Let $\mixcoef{k}$ denote the return process's $\beta$-mixing
coefficient at lag $k$. For each parity, Berbee's lemma couples the return
segments to independent copies. The total coupling-failure probability is at most
$C(T/\ell_{T,M})\mixcoef{c_{\rm b}\log T}\leq CT^{-A}$ for sufficiently
large $c_{\rm b}$ \citep{berbee1979random,yu1994rates}.
The coupling uses only return segments and is therefore common to all
deterministic network candidates.

We can now apply an independent-block concentration bound. For a given
candidate, let $S_j$ be the sum of its losses over coupled block $j$.
It satisfies
$0\leq S_j\leq C\ell_{T,M}$ and
$\E S_j^2\leq C\ell_{T,M}\E S_j$. Bernstein's inequality, applied to
each parity, and a union bound over the cover give the relative-error and
sampling terms. Restoring the omitted dates and the true variance adds
boundary and truncation errors. The sampling and truncation remainders cover these
errors and the return tails, proving \eqref{eq:apt-stationary-transfer}.

We now combine these bounds with the initial decomposition.
We first bound $Q_{\mathcal A}(f)$ by its sample loss, apply the empirical
comparison, and then bound the sample loss of $f_{\snet}$ by its stationary risk.
For sufficiently small $\eta$, absorbing the sampling remainders into
$\mathfrak E_{T,M}$ gives, on the common high-probability event,
\begin{equation}
 Q_{\mathcal A}(f)
 \leq(1+C\eta)Q_{\mathcal A}(f_{\snet})
 +C_\eta\{\mathfrak E_{T,M}+\delta_M^{\rm tr}\}.
 \label{eq:apt-population-oracle}
\end{equation}
\smallskip
\noindent\textbf{Recursion-map risk.}\par\nobreak
We first decompose the function-estimation error. Let
$\widehat\sigma_{t,m}^{2\prime}$ denote the variance path generated on the
independent copy by the estimated initial value and the first $m$ fitted
networks. Write
$\Delta_{T,m}^{\rm stat}=\E|\widehat\sigma_{0,m}^{2\prime}-(\sigma_0')^2|^2$,
where the expectation is over both the copy and the estimation sample.
Adding and subtracting $\widehat\sigma_{0,M}^{2\prime}$ gives
\begin{align*}
 \E\{\widehat f_M(X_{0,*}')-\ftrue(X_{0,*}')\}^2
 &\leq2\Delta_{T,M}^{\rm stat}
 +2\E|\widehat f_M(X_{0,*}')-\widehat\sigma_{0,M}^{2\prime}|^2\\
 &\leq2\Delta_{T,M}^{\rm stat}
 +2q(B')^2\Delta_{T,M-1}^{\rm stat}.
\end{align*}
The second inequality uses the variance-input Lipschitz bound for
$\widehat f_M$ and stationarity. Only the variance inputs differ, and the
second term is absent when $q=0$. It therefore suffices to bound the
stationary variance errors at iterations $M$ and $M-1$.

To control these errors, we substitute the earlier comparison-network bound
into \eqref{eq:apt-population-oracle}. The coefficient on the preceding
variance error is
$(1+C\eta)(1+\tau)\Dcon^2$. We choose $\tau$ and then $\eta$
sufficiently small to make this coefficient at most $\rho$.
Because the comparison holds uniformly, it also applies to the estimated initial
value, preceding networks, and current empirical minimizer. Taking expectations and using
bounded losses on the exceptional event gives
\begin{equation}
 \Delta_{T,m}^{\rm stat}
 \leq\rho\Delta_{T,m-1}^{\rm stat}
 +C\left\{\|f_{\snet}-\ftrue\|_{\infty,\mathcal X_T}^2
 +\mathfrak E_{T,M}+\delta_M^{\rm tr}\right\}.
 \label{eq:apt-stage-recursion}
\end{equation}
Iterating this inequality gives
\[
 \Delta_{T,m}^{\rm stat}\leq C\rho^m+
 \frac{C}{1-\rho}\left\{\|f_{\snet}-\ftrue\|_{\infty,\mathcal X_T}^2
 +\mathfrak E_{T,M}+\delta_M^{\rm tr}\right\},
\]
with no initialization term when $\qstar=0$. For $\qstar=0<q$, we take
$M\geq2$ to apply the stagewise bound also to $\Delta_{T,M-1}^{\rm stat}$.

The approximation error is at most
$C(\log T)^c\snet^{-2\hol/(\fdim)}$ by Proposition~\ref{prop:network-up}.
Under the choices in Theorem~\ref{thm:main-upfront},
$R_M,\ell_{T,M}=O(\log T)$, and both the approximation error and
$\mathfrak E_{T,M}$ are at most $Cr_T$. A sufficiently large constant in
$M\asymp\log T$ gives
$\rho^{M-1}+\delta_M^{\rm tr}\leq Cr_T$.
Substitution into the error decomposition yields
$\E\{\widehat f_M(X_{0,*}')-\ftrue(X_{0,*}')\}^2\leq Cr_T$,
proving the first assertion.

\smallskip
\noindent\textbf{Future forecast risk.}\par\nobreak
We transfer the preceding function-estimation bound to forecasts by showing
that, for $k\geq1$ and some $\lambda_{\mathrm F}\in(0,1)$,
\begin{align*}
 \E\{\widehat\sigma^2_{T+k\mid T}-\volsq_{T+k}\}^2
 &\leq C\E\{\widehat f_M(X_{0,*}')-\ftrue(X_{0,*}')\}^2
 +C\lambda_{\mathrm F}^k\\
 &\leq Cr_T+C\lambda_{\mathrm F}^k.
\end{align*}
The additional term accounts for initialization and dependence on the
estimation sample. We first apply $\widehat f_M$ to the independent
stationary copy, starting at date $u$ from a deterministic variance state
in $[\Bmin,\Bmax]^q$. For $r>u$, let $e_r$ be the absolute difference
between this forecast and $(\sigma_r')^2$. The network's variance-input
contraction gives, when $q\geq1$,
\[
 e_r\leq B'\max_{1\leq j\leq q}e_{r-j}
 +|\widehat f_M(X_{r,*}')-\ftrue(X_{r,*}')|.
\]
On this independent copy, stationarity gives the same expected squared
function error at every date. Since $B'<1$, grouping dates into blocks of
length $q$ and applying Young's inequality yields, for $L\geq1$,
\[
 \E e_{u+L}^2
 \leq C\E\{\widehat f_M(X_{0,*}')-\ftrue(X_{0,*}')\}^2
 +C\rho_{\mathrm F}^{\lfloor L/q\rfloor},
\]
for some $\rho_{\mathrm F}\in(0,1)$. When $q=0$, only the
function-estimation term remains.

To transfer this bound to the actual future path, we construct an auxiliary
forecast using only a later return segment. For $k>p+2$, we set $t=T+k$,
$L=\lfloor(k-p)/2\rfloor$, and $u=t-L$. The auxiliary forecast starts at
$u$ from the same deterministic variance state and uses the observed
returns through $t-1$. Contraction of $\widehat f_M$ bounds its squared
difference from the original forecast by $C(B')^{2\lfloor L/q\rfloor}$
when $q\geq1$. When $q=0$, the forecasts agree. The required return
segment starts at $u-p+1$, leaving a gap of at least $k-L-p$ dates from
the estimation sample. Berbee's lemma couples this segment to an independent
stationary copy with failure probability at most $\mixcoef{k-L-p}$.
On the coupling event, the auxiliary forecasts on the observed and independent
paths agree. The corresponding
true variances may still differ because their earlier histories differ.
Contraction of $\ftrue$ bounds this squared difference by
$C\Dcon^{2\lfloor L/\qstar\rfloor}$ when $\qstar\geq1$, and it is zero
when $\qstar=0$. Bounded variances limit the contribution of coupling
failure to $C\mixcoef{k-L-p}$.

Both $L$ and the separation gap $k-L-p$ are of order $k$. Since $B'<1$,
$\Dcon<1$, and the mixing coefficients decay geometrically, the
initialization and dependence terms are bounded by $C\lambda_{\mathrm F}^k$.
Combining these terms with the independent-path bound gives the desired
inequality. Increasing $C$ covers the finitely many smaller values of $k$
and proves the second bound in \eqref{eq:apt-population-bounds}.

\smallskip
\noindent\textbf{Expected excess QLIKE loss.}\par\nobreak
For the loss comparison, let $v=\volsq_{T+k}$ and
$w=\widehat\sigma^2_{T+k\mid T}$. Conditional
on $\filt{T+k-1}$, the expected excess QLIKE of $w$ over the oracle
forecast $v$ equals $v/w-\log(v/w)-1$. Taylor's theorem on
$[\Bmin,\Bmax]^2$ gives
$c_{\mathrm Q}(v-w)^2\leq v/w-\log(v/w)-1\leq
C_{\mathrm Q}(v-w)^2$.  Taking expectations proves the result.
\end{proof}

\section{Proofs of the remaining population and forecasting results}
\label{app:apt-rates}

The recursion-map bound in \eqref{eq:apt-population-bounds} gives
Corollary~\ref{cor:func-recovery}. Averaging the forecast bound over
$T+N+1,\ldots,T+N+H$ gives Theorem~\ref{thm:oos-upfront}.
Its $H=1$ case and the QLIKE comparison in
Lemma~\ref{lem:population-transfer} at $k=N+1$ give
Corollary~\ref{cor:qlike-upfront}.

\section{Additional simulation results}
\label{app:simulation-results}

\begin{figure}[!htbp]
\centering
\linespread{1}\selectfont
\includegraphics[width=0.94\textwidth]{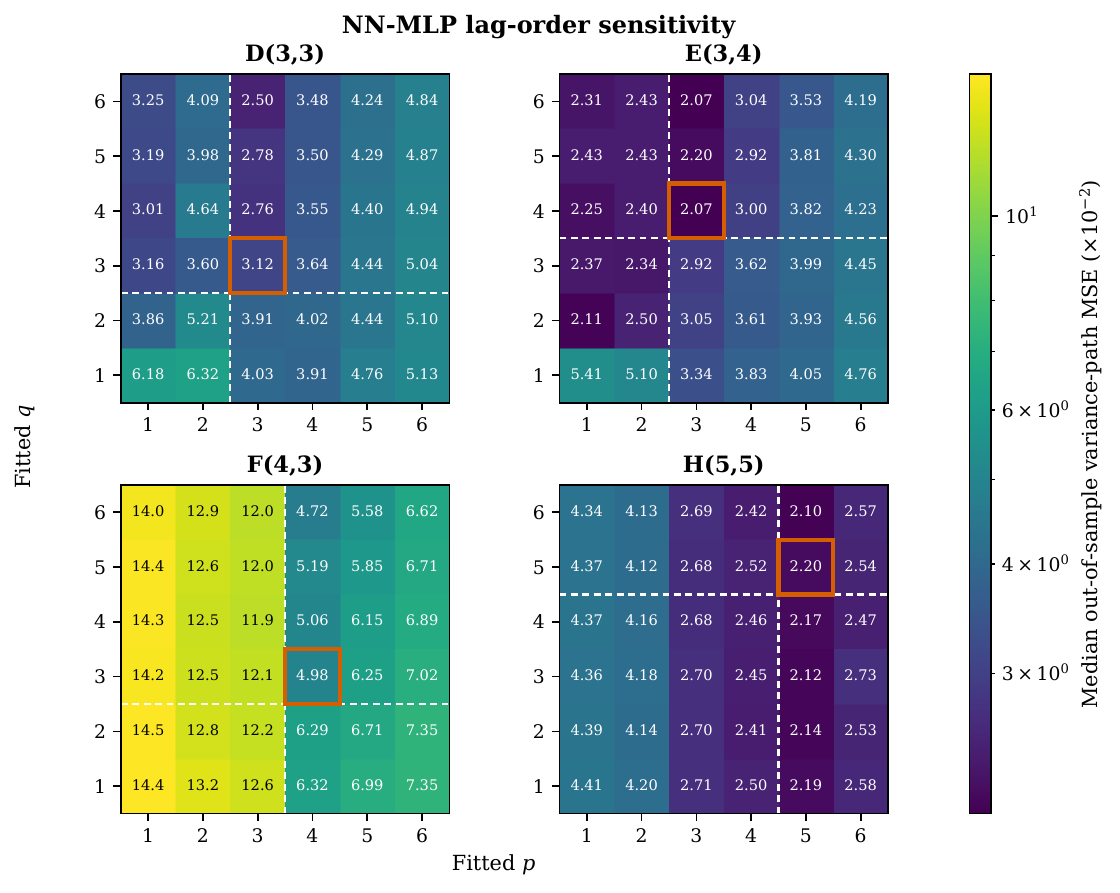}
\caption{NN-MLP lag-order sensitivity under nonlinear
stationary dynamics. Each cell reports the median out-of-sample variance-path
MSE over 20 replications, scaled by 100. The orange box marks the structural
lag orders, and the dashed lines mark the corresponding return- and
variance-lag boundaries.}
\label{fig:lag-order-defh-nn}
\end{figure}

Figure~\ref{fig:lag-order-defh-nn} extends the lag-order
comparison to NN-MLP under designs D, E, F, and H. Sensitivity to omitted lags
varies across the four designs, whereas additional lags generally change the
forecast error more gradually.

\FloatBarrier

\end{document}